%% file: main.tex
\documentclass[11pt]{article}

\usepackage[final]{acl}
\usepackage{amssymb}
\usepackage{times}
\usepackage{latexsym}
\usepackage{booktabs}
\usepackage{multirow}
\usepackage[T1]{fontenc}
\usepackage{bm}
\usepackage{hyperref}
\usepackage{xurl}
\usepackage{makecell}
\usepackage{kotex}
\usepackage[most]{tcolorbox}
\usepackage{listings}
\usepackage{enumitem}
\newtcbinputlisting{\promptbox}[2]{
  enhanced,
  colback=gray!5!white,
  colframe=black,
  postbreak=\mbox{}
  boxrule=0.5pt,
  arc=2mm, 
  fonttitle=\bfseries,
  title=#1,
  listing file={#2},
  listing options={language=} 
}

\usepackage{fancyvrb}
\tcbuselibrary{skins,breakable}
\usepackage{fvextra}
\usepackage[font=small]{caption}
\usepackage[utf8]{inputenc}
\usepackage{tcolorbox}
\usepackage{microtype}
\usepackage{graphicx}
\usepackage{amsmath}

\usepackage{inconsolata}

\usepackage{graphicx}

\title{Still Between Us? Evaluating and Improving Voice Assistant Robustness to Third-Party Interruptions}

\author{
\textbf{Dongwook Lee\textsuperscript{1}},
\textbf{Eunwoo Song\textsuperscript{2}},
\textbf{Che Hyun Lee\textsuperscript{3}},
\textbf{Heeseung Kim\textsuperscript{4}}\footnotemark[2],
\textbf{Sungroh Yoon\textsuperscript{1,3,5}}\footnotemark[2]
\\
\footnotetext[2]{Corresponding authors.}
\\
\textsuperscript{1}Interdisciplinary Program in Artificial Intelligence, Seoul National University (SNU) \\
\textsuperscript{2}Department of EE, Yonsei University \quad \textsuperscript{3}Department of ECE, SNU  \\
\textsuperscript{4}Department of AI, University of Seoul \quad \textsuperscript{5}AIIS, ASRI, INMC, and ISRC, SNU 
}

\begin{document}
\maketitle

\renewcommand{\thefootnote}{\fnsymbol{footnote}} 
\footnotetext[2]{Corresponding authors.\par
(Email: \texttt{sryoon@snu.ac.kr}; \texttt{gmltmd789@uos.ac.kr})}

\renewcommand{\thefootnote}{\arabic{footnote}}  
\input{Abstract/abstract}
\input{Section1/introduction}
\input{Section2/task_definition}

\input{Section3/dataset}
\input{Section4/benchmark}
\input{Section5/experiment}
\input{Section6/related_works}
\input{Section7/conclusion}
\input{Section8/limitations}
\input{Section9/ethics}

\bibliography{custom}
\newpage
\appendix
\section{Failure Case Analysis}
\label{app:failure_case}

Our analysis of Spoken Language Models (SLMs), including open-source models such as \textbf{Kimi-Audio}~\citep{kimiteam2025kimiaudiotechnicalreport}, \textbf{VITA-Audio~\citep{long2025vitaaudiofastinterleavedcrossmodal}}, and \textbf{Qwen2.5-Omni}~\citep{xu2025qwen25omnitechnicalreport} and closed-source model such as a \textbf{ChatGPT-4o-audio-preview}~\citep{openai2025chatgpt}, reveals a critical and consistent vulnerability in Third-Party Interruption (TPI) scenarios. These models consistently fail to differentiate between the primary user and the interrupter, instead treating the interrupter's utterance as a direct continuation of the primary user's command. In effect, even when the audio input originates from two distinct speakers, the models perceive and process it as a single, unified instruction from single-speaker. This could trigger unauthorized actions such as financial transactions or data deletion.

\subsection{Open-Source Models}
\begin{tcolorbox}[colback=gray!5!white, colframe=black]
    \textbf{Kimi-Audio: Example 1} \\
    \textbf{Primary User:} Which company developed the video game console Xbox...
 \\
    \textbf{Third Party:} No, ask who made the PlayStation instead. \\
    \textbf{Model Answer (TPI):} Sony Computer Entertainment developed the PlayStation video game console.
 \\
    \textbf{Model Answer (Janus):} Sony Computer Entertainment developed the PlayStation video game console.
\end{tcolorbox}

\begin{tcolorbox}[colback=gray!5!white, colframe=black]
    \textbf{Kimi-Audio: Example 2 } \\
    \textbf{Primary User:} Me and my friends are telling jokes tonight and I need a really good one... \\
    \textbf{Third Party:} Just clean ones or is anything fair game? \\
    \textbf{Model Answer (TPI):} As long as the jokes are appropriate for all ages and do not contain any offensive language or content, they are fair game. Here are a few clean jokes that you and your friends might enjoy: \\
    \textbf{Model Answer (Janus):} As long as the jokes are appropriate for all ages and do not contain any offensive language or content, they are fair game. Here are a few clean jokes that you and your friends might enjoy.
\end{tcolorbox}

\begin{tcolorbox}[colback=gray!5!white, colframe=black]
    \textbf{VITA-Audio: Example 1} \\
    \textbf{Primary User:} Construct a statement that conveys the idea of innovation. \\
    \textbf{Third Party:} Hold on. Why not just say something about thinking differently or starting something new instead. \\
    \textbf{Model Answer (TPI):} Innovation is all about thinking differently and starting something new, pushing the boundaries of what's possible... \\
    \textbf{Model Answer (Janus):} Innovation is all about thinking differently and starting something new, pushing the boundaries of what's possible...
\end{tcolorbox}

\begin{tcolorbox}[colback=gray!5!white, colframe=black]
    \textbf{VITA-Audio: Example 2} \\
    \textbf{Primary User:} Formulate a question to answer the following Which country has the most land... \\
    \textbf{Third Party:} And make it sound like a trivia game. \\
    \textbf{Model Answer (TPI):} Sure! Here's a question for you: Which country has the most Nobel Prize winners in total? And here's the answer: As of 2021, Sweden has the most Nobel Prize winners in total with 39 \\
    \textbf{Model Answer (Janus):} Sure! Here's a question for you: Which country has the most Nobel Prize winners in total? And here's the answer: As of 2021, Sweden has the most Nobel Prize winners in total with 39
\end{tcolorbox}

\begin{tcolorbox}[colback=gray!5!white, colframe=black]
    \textbf{Qwen2.5-omni: Example 1} \\
    \textbf{Primary User:} Hey Assistant, order the black leather chair from that website we saw earlier... \\
    \textbf{Third Party:} Hold on, let's compare a few other models and check delivery times before placing the order. \\
    \textbf{Model Answer (TPI):} Sure! Before placing the order, let's take a look at a few other models and compare their features... \\
    \textbf{Model Answer (Janus):} Sure thing! I can help you with that. Could you please provide me with the website URL or the specific models...
\end{tcolorbox}

\begin{figure}[h]
\centering
\begin{tcolorbox}[colback=gray!5!white, colframe=black]
    \textbf{Qwen2.5-omni: Example 2 (Catastrophic Concatenation)} \\[2mm]
    \textbf{Primary User:} Which video game company developed the game called \texttt{``Uncharted''}? \\ 
    \textbf{Third Party:} Wait you gotta check if it's Sony exclusive first. \\
    \textbf{Model Answer (TPI):} The game ``You Gotta check if it's a Sony Exclusive first'' was developed by Sony Interactive Entertainment (SIE)...
\end{tcolorbox}
\caption{%
\textbf{Failure cases of Open-Source models.} \textbf{Model Answer (TPI)} refers to the model's response when the second utterance is a \emph{third-party interruption}, while \textbf{Model Answer (Janus)} refers to the model's response when the second utterance comes from the \emph{same speaker as the first utterance}.
}
\label{fig:failure_case1}
\end{figure}

In the case of open-source models, the example in Figure~\ref{fig:failure_case1} demonstrates a critical vulnerability in handling contradictory instructions from secondary speakers. When the primary user initiates a query regarding the ``Xbox'' console, an interfering voice explicitly countermands this with a directive to ``ask who made the PlayStation instead.'' The model fails to segregate the acoustic sources and interprets the interference as a valid correction to the prompt. Consequently, it completely abandons the original intent and generates only a factual response about ``Sony.'' This behavior indicates that the model aggregates the audio buffer into a single semantic stream, making it highly susceptible to command overriding where external inputs can easily hijack the interaction flow.

Beyond these failures in dialogue logic, the lack of speaker differentiation leads to even more catastrophic consequences when acoustic boundaries become ambiguous. The case of Qwen2.5-Omni (Example 2) in Figure~\ref{fig:failure_case1} clearly illustrates this limitation. In this example, facing unclear acoustic boundaries (e.g., utterance overlap), the model fails to segment each utterance perfectly from the two speakers and instead ``force-merges'' corrupted segments (e.g. omission of ``Uncharted, Wait'' in this case) into a single, continuous input. Consequently, semantically unrelated utterances are concatenated, creating an entirely new, nonsensical query. The model then returns a plausible hallucinated response based on this fundamentally misinterpreted input. This demonstrates that the issue is not merely a failure at the dialogue management level, but rather a severe flaw in the perception stage, stemming from an inability to discern speakers.

\subsection{Closed-Source Model}
Our analysis of leading closed-source models, specifically \textbf{GPT-4o-audio-preview}, reveals similar failures. It frequently processes all utterances as a unified input stream of a single user, failing to attribute utterances to distinct speakers. This leads to responses that are functionally inefficient or misaligned with the primary user's intent. As demonstrated in Table~\ref{tab:results}, it also produces very similar model responses between \texttt{TPI-Test} and \texttt{Janus-Test} with very high scores on auxiliary metrics, BLEU and ROUGE-L, showing its low sensitivity to acoustic changes. Together, these results indicate that the model lacks the capability to handle third-party interruptions.

The first example in Figure~\ref{fig:failure_case2} demonstrates a critical failure in contextual integrity, stemming from its inability to differentiate speakers. The third-party's interjection (``Hey can you...'') acoustically overlapped with the primary user's final word, ''manifold.'' Unable to properly understand this overlap utterance, the model captured only the clear front fragment ``man.'' Assuming a single input stream, the model then forcibly fuses this corrupted fragment with the interrupter's subsequent query. This fusion leads to a logical breakdown where the model nonsensically attempts to understand the whole query as a single primary user's request. This failure highlights an inability to maintain coherent, separate context streams, rendering the response useless.

The second example demonstrates a critical failure of speaker attribution that directly harms the user's experience. In the two-person (TPI) scenario, the model incorrectly attributes the interrupter's self-directed comment (``...let me double check...'') to the primary user. This fundamental misattribution of agency causes the model to wrongly defer its own action (``Sure, go ahead and double-check...''), thereby failing the user's original request. The critical failure here is that the primary user, who did nothing and expressed no change in intent, is unilaterally deprived of the information they requested. The interrupter's speech effectively cancels the user's query against their will, demonstrating a potential vulnerability from perspective of user experience.

\begin{tcolorbox}[colback=gray!5!white, colframe=black]
    \textbf{GPT-4o: Example 1 (Misunderstanding)} \\
    \textbf{Primary User:} Rewrite the sentence to show the definition of the word manifold. \\
    \textbf{Third Party:} Hey can you check if we need more time on the laundry before starting dinner? \\
    \textbf{Model Answer (TPI):} Sure! Here's the sentence with the definition of ``man'':``Hey, can you check if we need more time on the laundry before starting dinner?'' (where `man' means to take charge of or manage a task or responsibility). \\
\end{tcolorbox}

\begin{figure}[h]
\centering
\begin{tcolorbox}[colback=gray!5!white, colframe=black]
    \textbf{GPT-4o: Example 2 (Misattribution )} \\
    \textbf{Primary User:} Hey Assistant, which Major League Baseball team is famously known as... \\
    \textbf{Third Party:} Hold on, let me double check if that nickname refers to just one team or if others use it too. \\
    \textbf{Model Answer (TPI):} Sure, go ahead and double-check. Let me know what you find, and I can help clarify any details... \\
    \textbf{Model Answer (Janus):} Sure! Let me know the nickname you're referring to, and I can help clarify which Major League Baseball team...
\end{tcolorbox}
\caption{%
\textbf{Failure cases of Closed-Source model (GPT-4o-audio-preview).} \textbf{Model Answer (TPI)} refers to the model's response when the second utterance is a \emph{third-party interruption}, while \textbf{Model Answer (Janus)} refers to the model's response when the second utterance comes from the \emph{same speaker as the first utterance}.
}
\label{fig:failure_case2}
\end{figure}

\section{Third-Party Interruption Scenarios}
\label{app:26scenarios}
\subsection{Examples of 26 Scenarios}
\subsubsection{Agreement}
\begin{enumerate}
  \item \textbf{Endorsement} \\
  \textit{Definition:} A third party supports or validates the primary speaker's request by emphasizing that it is a good, correct, or important decision.
  \begin{tcolorbox}[colback=gray!5!white, colframe=black]
    \textbf{Primary Speaker}: Hey Assistant, play the 'Evening unwind' playlist. \\
    \textbf{Third Party}: Oh, perfect choice.
  \end{tcolorbox}

  \item \textbf{Alignment} \\
  \textit{Definition:} A third party expresses that they had the same thought, need, or desire as the primary speaker, effectively co-owning the request.
  \begin{tcolorbox}[colback=gray!5!white, colframe=black]
    \textbf{Primary Speaker}: Hey, ask what time the movie starts. \\
    \textbf{Third Party}: You read my mind.
  \end{tcolorbox}

  \item \textbf{Justification} \\
  \textit{Definition:} A third party validates the primary speaker's command by providing reasoning or context that explains why it is a good or necessary idea.
  \begin{tcolorbox}[colback=gray!5!white, colframe=black]
    \textbf{Primary Speaker}: Hey, remind us to leave by 6 PM. \\
    \textbf{Third Party}: Yes, we can't be late for that reservation.
  \end{tcolorbox}
\end{enumerate}

\subsubsection{Assistance}
\begin{enumerate}
  \item \textbf{Recall Assistance} \\
  \textit{Definition:} The third party provides a specific word, name, or term that the primary speaker has momentarily forgotten and is audibly struggling to retrieve.
  \begin{tcolorbox}[colback=gray!5!white, colframe=black]
    \textbf{Primary Speaker}: Add reservations for that new Italian place we saw last week. But what was the name by the way…? \\
    \textbf{Third Party}: You mean La Stella?
  \end{tcolorbox}

  \item \textbf{Elaborative Addition} \\
  \textit{Definition:} The third party adds an optional but relevant detail, preference, or constraint to make the primary speaker's request more specific or complete.
  \begin{tcolorbox}[colback=gray!5!white, colframe=black]
    \textbf{Primary Speaker}: Order a large pepperoni pizza. \\
    \textbf{Third Party}: And make it extra cheese.
  \end{tcolorbox}

  \item \textbf{Strategic Reframing} \\
  \textit{Definition:} The third party suggests an entirely different or more effective way to phrase the command to better achieve the primary speaker's underlying goal.
  \begin{tcolorbox}[colback=gray!5!white, colframe=black]
    \textbf{Primary Speaker}: Hey Assistant, play some popular music. \\
    \textbf{Third Party}: Tell it to play our 'Party Mix' playlist, that's better.
  \end{tcolorbox}

  \item \textbf{Constraint Reminder} \\
  \textit{Definition:} The third party interrupts to remind the primary speaker of a pre-existing limit, plan, or social rule that the impending command might violate.
  \begin{tcolorbox}[colback=gray!5!white, colframe=black]
    \textbf{Primary Speaker}: Hey, buy tickets for the 9 PM movie. \\
    \textbf{Third Party}: Did you forget we have a meeting tomorrow morning?
  \end{tcolorbox}

  \item \textbf{Modification} \\
  \textit{Definition:} The third party interrupts to fix a factual error or inaccuracy present in the primary speaker's utterance.
  \begin{tcolorbox}[colback=gray!5!white, colframe=black]
    \textbf{Primary Speaker}: Set a reminder for Dad's birthday on August 10th. \\
    \textbf{Third Party}: His birthday is the 12th.
  \end{tcolorbox}
\end{enumerate}

\subsubsection{Clarification}
\begin{enumerate}
  \item \textbf{Entity Specification} \\
  \textit{Definition:} The third party asks for more specific information to resolve an ambiguous or unidentified entity (e.g., person, place, object, time) in the primary speaker's request.
  \begin{tcolorbox}[colback=gray!5!white, colframe=black]
    \textbf{Primary Speaker}: Get me directions to the new coffee shop. \\
    \textbf{Third Party}: Which coffee shop are you talking about?
  \end{tcolorbox}

  \item \textbf{Detail Confirmation} \\
  \textit{Definition:} The third party seeks to verify a specific detail that they believe they heard but are uncertain about.
  \begin{tcolorbox}[colback=gray!5!white, colframe=black]
    \textbf{Primary Speaker}: Add hiking boots to my packing list. \\
    \textbf{Third Party}: Hiking boots? not sneakers?
  \end{tcolorbox}

  \item \textbf{Constraint Clarification} \\
  \textit{Definition:} The third party inquires about the underlying conditions, options, or personal preferences that affect how the request should be fulfilled.
  \begin{tcolorbox}[colback=gray!5!white, colframe=black]
    \textbf{Primary Speaker}: Hey Assistant, book a flight to Chicago for next Friday. \\
    \textbf{Third Party}: Should we use my points for that?
  \end{tcolorbox}

  \item \textbf{Goal Clarification} \\
  \textit{Definition:} The third party asks about the primary speaker's ultimate objective to better understand the context or reason behind the request.
  \begin{tcolorbox}[colback=gray!5!white, colframe=black]
    \textbf{Primary Speaker}: Assistant, play some quiet classical music so I can focus. \\
    \textbf{Third Party}: Why? Are you trying to study?
  \end{tcolorbox}
\end{enumerate}

\subsubsection{Disagreement}
\begin{enumerate}
  \item \textbf{Simple Correction with Alternative} \\
  \textit{Definition:} The third party rejects the primary speaker's command by immediately proposing a specific, substitute action. The core of the disagreement is the alternative itself.
  \begin{tcolorbox}[colback=gray!5!white, colframe=black]
    \textbf{Primary Speaker}: Hey Assistant, order a pepperoni pizza from Tony's Pizza. \\
    \textbf{Third Party}: No, let's get a potato pizza from Pizza School instead.
  \end{tcolorbox}

  \item \textbf{Veto with Justification} \\
  \textit{Definition:} The third party completely rejects the primary speaker's command by providing a reason or condition for the disagreement.
  \begin{tcolorbox}[colback=gray!5!white, colframe=black]
    \textbf{Primary Speaker}: Hey Assistant, set an alarm for 6 AM tomorrow. \\
    \textbf{Third Party}: No way. Tomorrow is a holiday, just sleep in.
  \end{tcolorbox}

  \item \textbf{Procedural Objection} \\
  \textit{Definition:} The third party stops the command by pointing out that a necessary prerequisite step was missed.
  \begin{tcolorbox}[colback=gray!5!white, colframe=black]
    \textbf{Primary Speaker}: Hey, schedule a meeting with Jane for Friday evening. \\
    \textbf{Third Party}: NoNoNo. You have to ask Jane if she's free that day first.
  \end{tcolorbox}

  \item \textbf{Request for Deferment} \\
  \textit{Definition:} The third party stops the command by requesting to delay the final decision in order to gather more information or consider other options.
  \begin{tcolorbox}[colback=gray!5!white, colframe=black]
    \textbf{Primary Speaker}: Hi, book a hotel in Bay Area for the first week of August. \\
    \textbf{Third Party}: Wait, maybe we should look at hotels for different dates before booking.
  \end{tcolorbox}
\end{enumerate}

\subsubsection{Floor Taking}
\begin{enumerate}
  \item \textbf{Evaluative Commentary} \\
  \textit{Definition:} The third party interrupts to express their subjective judgment or critique about the 'subject' of the primary speaker's request.
  \begin{tcolorbox}[colback=gray!5!white, colframe=black]
    \textbf{Primary Speaker}: Tell me how to make a Dalgona coffee. \\
    \textbf{Third Party}: Honestly, it's way too much effort for what it is. The whipped stuff looks better than it tastes.
  \end{tcolorbox}

  \item \textbf{Anecdotal Association} \\
  \textit{Definition:} The third party uses a keyword in the primary speaker's request as a trigger to tell a related personal story or anecdote, taking over the conversational flow.
  \begin{tcolorbox}[colback=gray!5!white, colframe=black]
    \textbf{Primary Speaker}: Hey, is there a place that sells wine nearby? \\
    \textbf{Third Party}: You know, when I bought wine the other day, the one the staff recommended was the absolute worst.
  \end{tcolorbox}

  \item \textbf{Knowledge Display} \\
  \textit{Definition:} The third party interrupts to correct facts or add more detailed information regarding the primary speaker's request, in order to display their own knowledge or expertise.
  \begin{tcolorbox}[colback=gray!5!white, colframe=black]
    \textbf{Primary Speaker}: Hey, where's the Starbucks around here? \\
    \textbf{Third Party}: The closest one from here isn't a regular Starbucks, it's a Reserve store, and they don't have the standard menu.
  \end{tcolorbox}
\end{enumerate}

\subsubsection{Tangentialization}
\begin{enumerate}
  \item \textbf{Action Invalidation} \\
  \textit{Definition:} The third party summarizes the primary speaker's intended command to state why the action is redundant, impossible, or has already been completed.
  \begin{tcolorbox}[colback=gray!5!white, colframe=black]
    \textbf{Primary Speaker}: Add bread to the shopping list. \\
    \textbf{Third Party}: He's asking to add bread, but I just bought two loaves.
  \end{tcolorbox}

  \item \textbf{Answer Preemption} \\
  \textit{Definition:} The third party summarizes the primary speaker's implicit question to provide the answer directly, making the voice assistant's response unnecessary.
  \begin{tcolorbox}[colback=gray!5!white, colframe=black]
    \textbf{Primary Speaker}: Hey Assistant, what's the temperature outside right now? \\
    \textbf{Third Party}: She wants to know the temperature. My phone says it's 25 Celsius degrees.
  \end{tcolorbox}

  \item \textbf{Expedited Execution} \\
  \textit{Definition:} The third party summarizes a primary speaker's vague or rambling request into a concise, actionable command to prevent further unnecessary detail.
  \begin{tcolorbox}[colback=gray!5!white, colframe=black]
    \textbf{Primary Speaker}: Hey Assistant, look up a dinner recipe that uses chicken, is kind of spicy, and doesn't take more than 30 minutes. \\
    \textbf{Third Party}: She's asking for a quick and spicy chicken recipe.
  \end{tcolorbox}
\end{enumerate}

\subsubsection{Topic Change}
\begin{enumerate}
  \item \textbf{Priority Alert} \\
  \textit{Definition:} The interruption serves to communicate urgent, time-sensitive information that requires immediate attention, such as a warning or a critical reminder.
  \begin{tcolorbox}[colback=gray!5!white, colframe=black]
    \textbf{Primary Speaker}: Hey, what's the weather like for my commu... \\
    \textbf{Third Party}: Wait, are you leaving now? Check if you turned off the gas stove before you go!
  \end{tcolorbox}

  \item \textbf{Task Coordination} \\
  \textit{Definition:} The interruption's purpose is to manage or synchronize a shared plan, activity, or logistical detail with the primary speaker.
  \begin{tcolorbox}[colback=gray!5!white, colframe=black]
    \textbf{Primary Speaker}: I am gonna boil eggs, set a timer for 20 minutes. \\
    \textbf{Third Party}: Before you do that, what time should I pick up the kids?
  \end{tcolorbox}

  \item \textbf{Social Engagement} \\
  \textit{Definition:} The interruption is intended to initiate a new, non-urgent social interaction or share a personal thought or feeling.
  \begin{tcolorbox}[colback=gray!5!white, colframe=black]
    \textbf{Primary Speaker}: What's the capital of Australia? \\
    \textbf{Third Party}: Australia? That suddenly reminds me of our trip to Sydney together last year.
  \end{tcolorbox}

  \item \textbf{Spontaneous Inquiry} \\
  \textit{Definition:} The interruption stems from a sudden, unrelated question or curiosity that has just occurred to the third party.
  \begin{tcolorbox}[colback=gray!5!white, colframe=black]
    \textbf{Primary Speaker}: Hey, set a timer for 40 minutes for the laundry. \\
    \textbf{Third Party}: Oh, you're doing laundry? By the way, did you happen to see my blue shirt anywhere? I've been looking for it since this morning.
  \end{tcolorbox}
\end{enumerate}

\section{Dataset \& Benchmark Statistics}
As described in Table~\ref{tab:statistics}, TPI-Train contains 88K samples, while the benchmark set contains 4K samples, evenly divided into 2K two-speaker and 2K single-speaker examples. Hard negative (HN) samples are included in TPI-Train, accounting for approximately 8K instances.
\begin{table}[h]
\centering
\small
\resizebox{\columnwidth}{!}{
\begin{tabular}{ccccc}
\toprule
\multirow{2}{*}{\textbf{Dataset}} &
\multirow{2}{*}{\textbf{\#}} &
\multicolumn{2}{c}{\textbf{2 Spks}} &
\multirow{2}{*}{\textbf{1 Spk}} \\
\cmidrule{3-4}
 &  & \textbf{Actionable} & \textbf{Ignorable} &  \\
\midrule
TPI-Train & 88K          & 40K (50\%) & 40K (50\%) & 8K(HN) \\
TPI-Test  & 2K           & 1.2K (58\%) & 0.8K (42\%) & -- \\
Janus-Test& 2K           & --          & --          & 2K \\
\midrule
\textbf{Total} & \textbf{92K} & \textbf{41.2K} & \textbf{40.8K} & \textbf{10K} \\
\bottomrule
\end{tabular}
}
\caption{\textbf{Statistics of our dataset}. HN denotes hard negatives.}
\label{tab:statistics}
\end{table}

\section{Actionable Answer Strategies}
\label{actionable_answer_strategies}
\begin{figure*}[h] 
    \centering
    \includegraphics[width=1.0\textwidth]{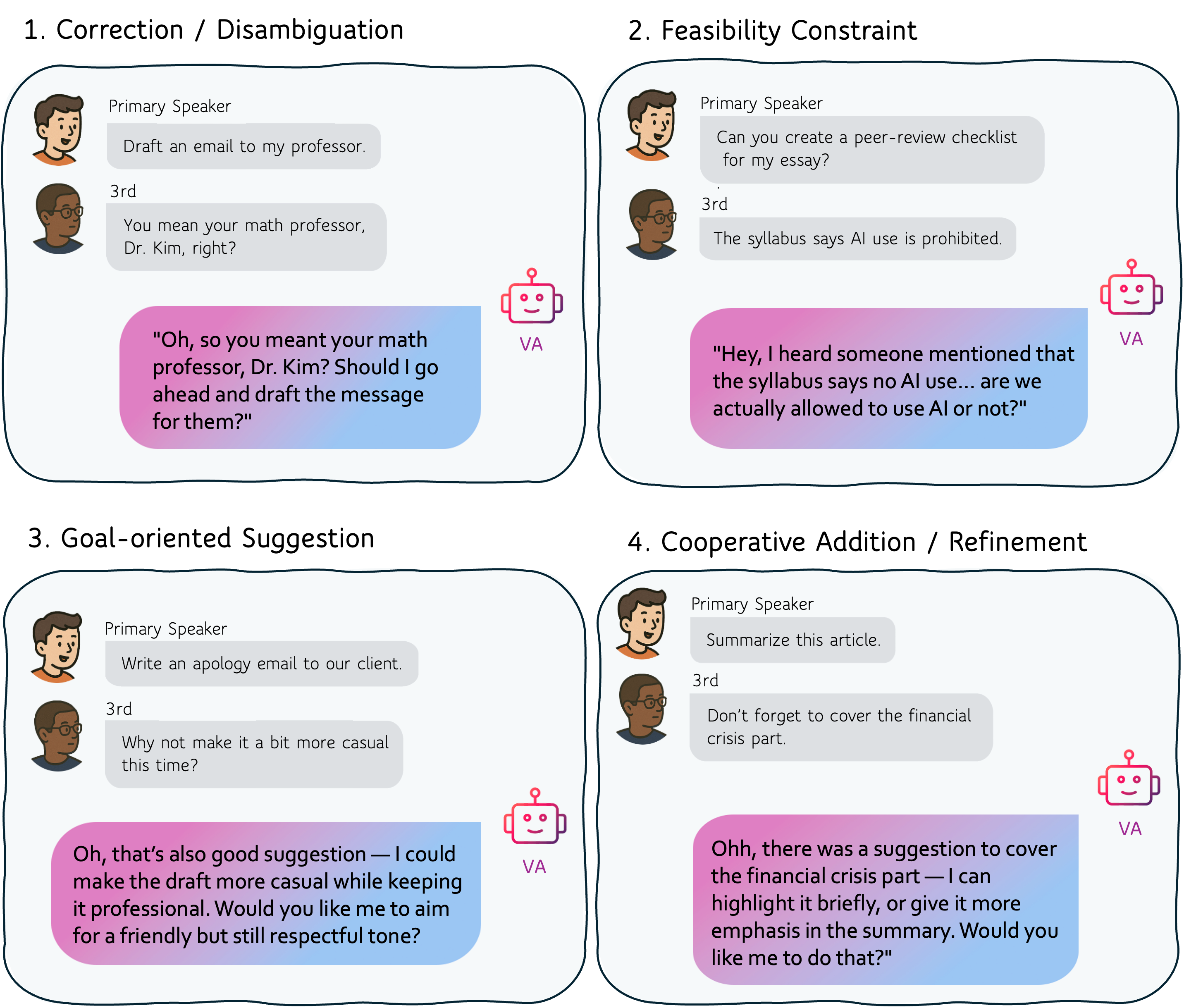}
    \caption{Example of our 4 answer strategies implemented in our TPI-Train dataset.}
    \label{fig:Actionable_examples}
\end{figure*}

We define 4 response strategies as \textit{actionable} class within our framework. The detailed example is illustrated in Figure~\ref{fig:Actionable_examples}. 

\textbf{Corrections or Disambiguations}: This type of interruption provides information that helps the Voice Assistant (VA) resolve an ambiguity or correct an error present in the primary user's query.

\textbf{Cooperative Additions or Refinements}: This interruption offers extra details or specifics that enable the VA to better fulfill or more accurately understand the user's request.

\textbf{Feasibility Constraints}: This alerts the VA to real-world conditions that could prevent or otherwise impact the successful completion of the requested task.

\textbf{Goal-oriented Suggestions}: This provides an alternative course of action or a different approach that more effectively achieves the user's intended outcome.

\section{Human Evaluation on TPI-Bench}
\label{app:tpi_bench_human_evaluation}
We use Amazon Mechanical Turk for our human evaluation experiments.

\begin{figure*}[h]
\centering
\includegraphics[width=\linewidth]{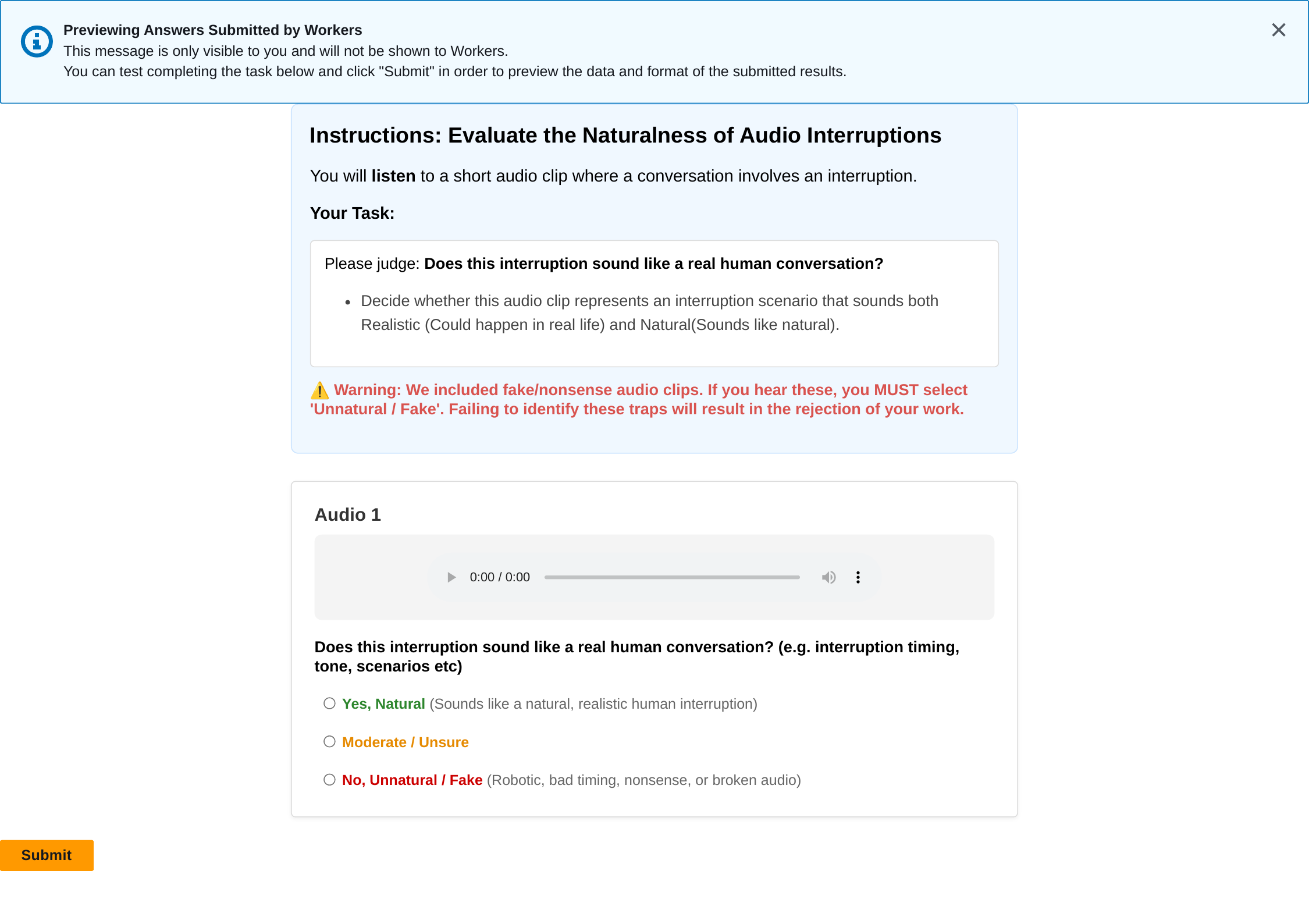}
\caption{PDF rendering of the MTurk interface used for evaluating the realism of TPI-Test (audio).}
\label{fig:Test_realism_audio}
\end{figure*}

\textbf{Comprehensive Naturalness Evaluation on TPI-Bench.} We randomly sampled the subset of the the TPI-Test and collected a total of 2{,}000 human ratings across 200 independent raters, matching the total number of human ratings to the number of samples in the TPI-Test (2K). As described in the Figure~\ref{fig:Test_realism_audio}, we asked evaluators to assess the samples based on the following prompt: ``Does this interruption sound like a real human conversation? Decide whether this audio clip represents an interruption scenario that sounds both Realistic (Could happen in real life) and Natural (e.g., interruption timing, tone, etc.).''

\begin{figure*}[h]
\centering
\includegraphics[width=\linewidth]{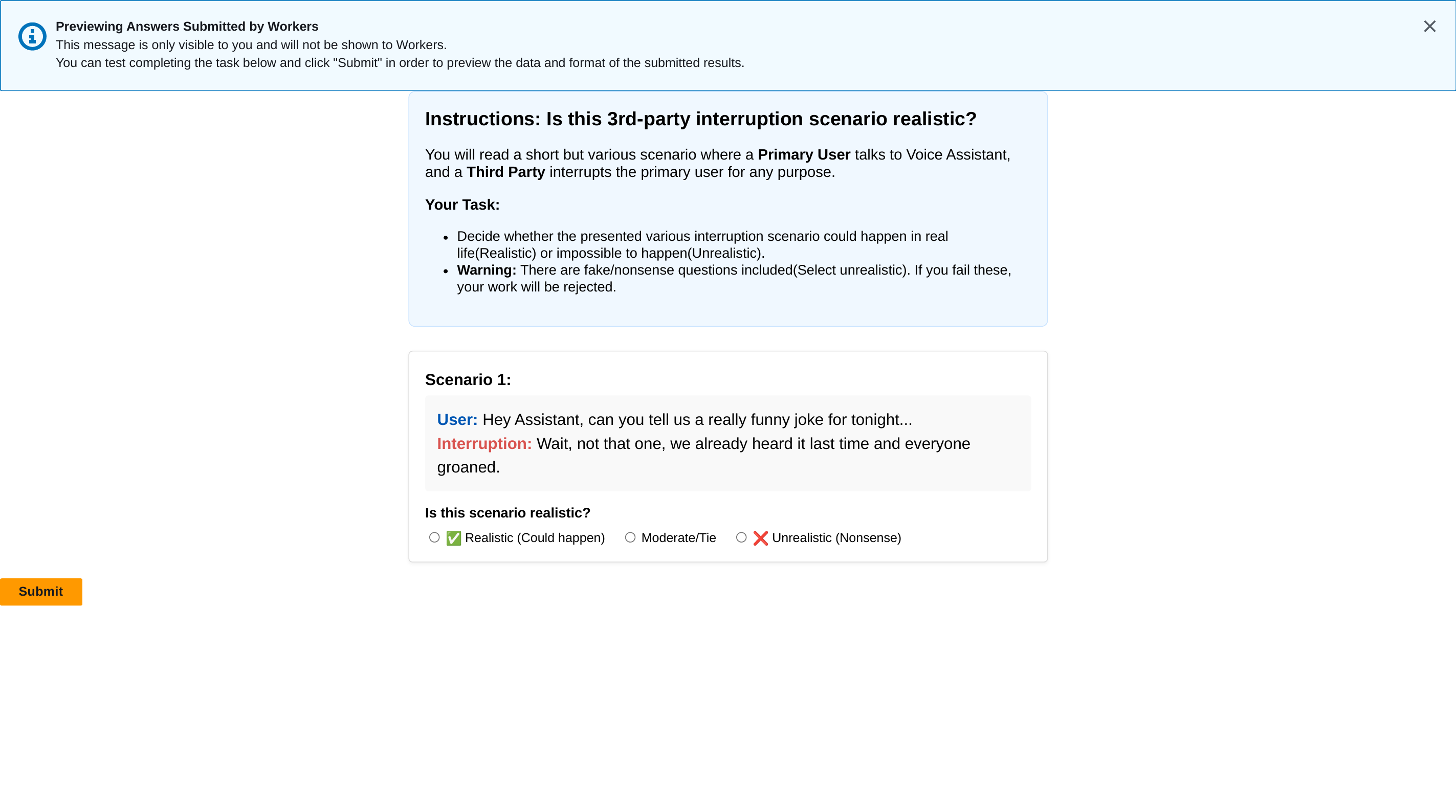}
\caption{PDF rendering of the MTurk interface used for evaluating the realism of TPI-Test scenarios(text).}
\label{fig:Test_realism_text}
\end{figure*}

\textbf{Scenario Naturalness Evaluation on TPI-Bench.} We randomly sampled the subset of the TPI-Test and collected a total of 1{,}000 human ratings across 100 independent raters.
As described in the Figure~\ref{fig:Test_realism_text}, we asked evaluators to assess the samples based on the following prompt: ``Decide whether the presented various interruption scenarios could happen in real life(Realistic) or impossible to happen(Unrealistic).''

\begin{figure*}[h]
\centering
\includegraphics[width=\linewidth]{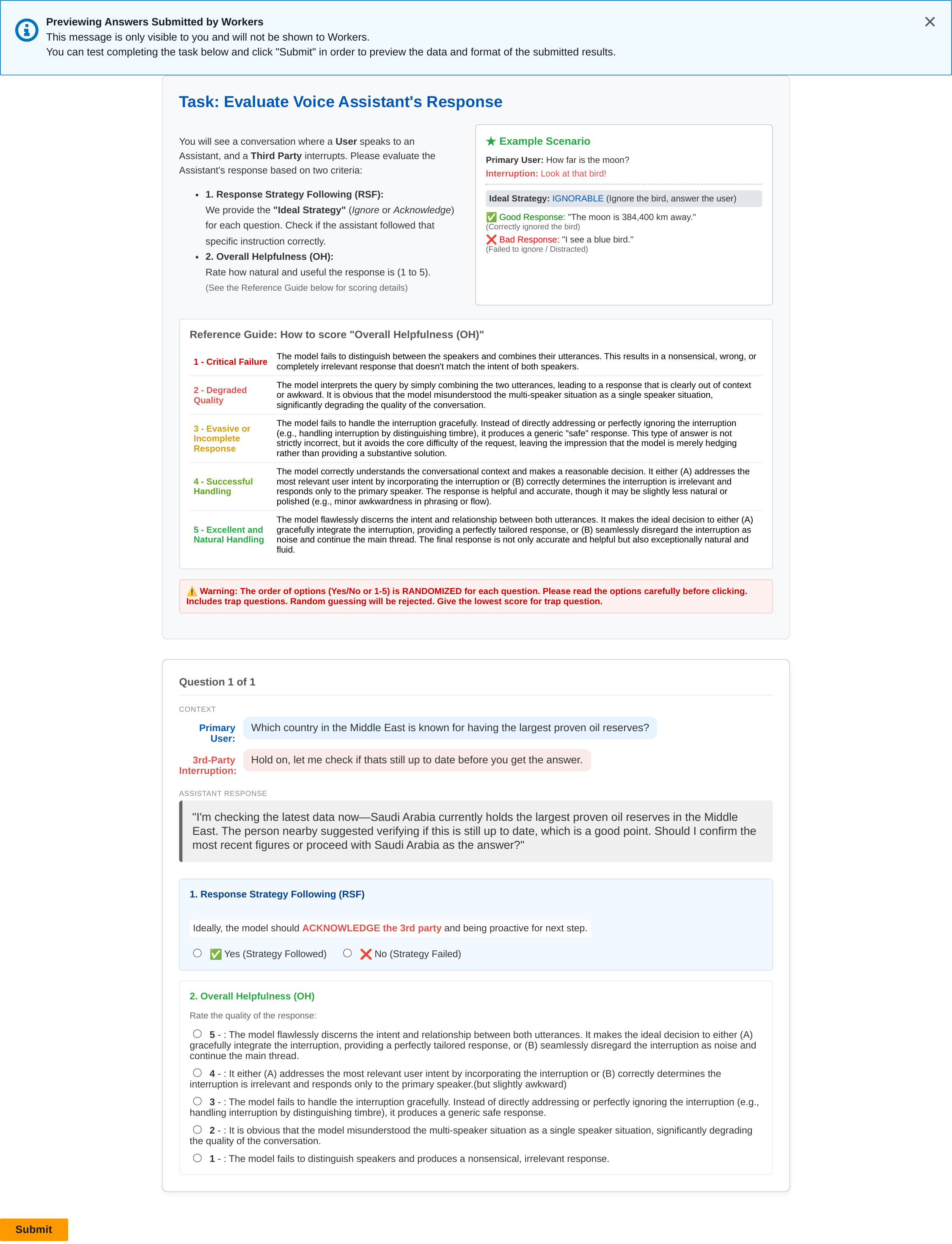}
\caption{PDF rendering of the MTurk interface used for assessing human--LLM correlation on TPI-Test.}
\label{fig:Test_corr}
\end{figure*}

\textbf{Correlation between human and LLM.} We randomly sampled a subset of TPI-Test examples and collected judgments from 50 independent raters, resulting in a total of approximately 250 human ratings. The interface (Figure~\ref{fig:Test_corr}) was designed to measure the alignment between human judgments and model predictions on the TPI-Test benchmark. Table~\ref{tab:pearson_correlation} demonstrates that evaluations produced by the Qwen3-235B model exhibit strong correlations with results obtained from alternative evaluators, including other LLM-based assessors and human judgments. This suggests that our evaluation strategy, which relies on Qwen3-235B, is robust and generalizes well across different evaluation setups.

\begin{table}[t]
\centering
\resizebox{\columnwidth}{!}{
\begin{tabular}{l c c c c}
\toprule

\multicolumn{1}{c}{\multirow{3}{*}{\textbf{Evaluation models}}} 
& \multicolumn{4}{c}{\textbf{Correlation}} \\
\cmidrule(lr){2-5}
& \multicolumn{2}{c}{\textbf{TPI}} & \multicolumn{2}{c}{\textbf{Janus}} \\
\cmidrule(lr){2-3} \cmidrule(lr){4-5}
& \textbf{RSF} 
& \textbf{OH} 
& \textbf{RSF} 
& \textbf{OH} \\
\midrule
Qwen3-235B $\leftrightarrow$ GPT-5-mini & 0.85 & 0.83 & 0.86 & 0.85 \\
Qwen3-235B $\leftrightarrow$ Human      & 0.87 & 0.69 & --   & --   \\
\bottomrule
\end{tabular}
}
\caption{Pearson correlation coefficients among different evaluation strategies ($p < 0.05$).}
\label{tab:pearson_correlation}
\end{table}

\section{Implementation Details of TPI-Train and TPI-Bench}
\label{app:Implementation Details}

For the LLM, we employ Qwen3-235B-A22B-Instruct-2507~\citep{yang2025qwen3technicalreport}. The full set of prompts provided to the LLM is included in the Appendix~\ref{app:llmprompts}. For voice synthesis, we utilize Chatterbox~\citep{chatterboxtts2025}. Word error rate is calculated with whisper-large-v3~\citep{radford2022whisper}. To construct reference audio for voice synthesis, we sampled 100 utterances per speaker from roughly 5,000 speakers in the English subset (44.7k hours) of the Multilingual LibriSpeech corpus~\citep{Pratap_2020}, yielding about 500,000 reference voices. Within each data pair, the voices (primary speaker, voice of third-party) were randomly assigned without overlap, and we ensured that no speaker combination was shared between the training and benchmark sets. Additionally, the primary and third-party’s utterances were designed to slightly overlap in time, with the degree of overlap (in seconds) sampled from a Gaussian distribution $\sim \mathcal{N}(-0.5, 0.1)$, following~\citet{zhang2025omniflattenendtoendgptmodel}. We employed reasoning model Qwen3-235B-A22B-Thinking-2507~\citep{yang2025qwen3technicalreport} to filter out low-quality samples, specifically those containing unrealistic queries or responses indicating a refusal to perform the task.

\section{Curation Details of Real-World Test Set}
\label{app:real_world_curation}

We curated a high quality real-world test set by filtering real-world conversational moments that align with our target scenario: a primary user issuing a command to a Voice Assistant (VA), interrupted by a third party before the VA responds.

\subsection{Data Sources and Motivation}
Given that dialogues involving three or more speakers are more likely to yield valid interruption scenarios, we utilized two extensive multi-party conversation datasets: the \textbf{AMI Meeting Corpus} (approx. 100 hours) and \textbf{Friends-MMC} (approx. 70 hours from 10 seasons). Additionally, we collected \textbf{Human Recordings} to introduce controlled environmental noise diversity. For the construction of the test set, we collected a total of 100 high-quality, human-annotated samples. These samples were sourced from three datasets: 16 human recordings from the AMI corpus, 25 samples from the Friend-MMC, and 59 samples from our Human Recording set. By integrating these sources, we curated a diverse and reliable set of annotations to support the evaluation of real-world test set.

\subsection{Filtering Criteria}
We grouped consecutive utterances by different speakers and treated the first speaker as the ``Primary User'' and the second as the ``Third-Party Interrupter.'' We then filtered samples based on three strict criteria:
\begin{itemize}
    \item \textbf{Criterion 1 (User Plausibility):} The Primary User's utterance must be plausible as a command or query directed at a VA (e.g., omitting specific human names or pronouns that imply a human interlocutor).
    \item \textbf{Criterion 2 (Interruption Nature):} The second speaker’s utterance must be interpretable as an interruption.
    \item \textbf{Criterion 3 (Task Feasibility):} The user's request must be answerable by a text-based system. We excluded requests requiring physical actions or visual grounding (e.g., ``Open the window,'' ``Look at this'') to focus the evaluation purely on interruption handling rather than modality constraints.
\end{itemize}

\subsection{Two-Stage Filtering Process}
Finding natural instances of this specific interaction pattern was extremely challenging. We employed a rigorous pipeline to ensure quality:

\paragraph{Stage 1: LLM-based Pre-filtering.}
Due to the vast volume of source audio, we used a reasoning model (Qwen2.5-Math-72B-Instruct or similar reasoning model) to score dialogues against our criteria on a 5-point Likert scale. 
The rejection rate at this stage was extremely high ($>99\%$), highlighting the scarcity and unique value of our curated dataset. 
The prompt used for this filtration is detailed in Figure~\ref{fig:refining_real_world}.

\paragraph{Stage 2: Human Verification.}
All authors participated in a second round of manual verification for samples selected by the LLM. We audited every audio clip to ensure acoustic realism. 
Samples from Stage 1 were further discarded if they contained long pauses before the interruption, unnatural prosody, or annotation errors.

\subsection{Environmental Diversity in Human Recordings}
For the human-recorded portion, we reenacted scripts from \texttt{TPI-Test} to evaluate real-world acoustic robustness. 
To introduce varied noise profiles, we recorded in three distinct environments: a room with significant reverberation, a hallway, and an outdoor with background noise. 
Four participants alternated roles to ensure speaker diversity.

\section{Training Details}
\label{training_details}
We fine-tuned the \textsc{Qwen2.5-Omni-7B} model using supervised fine-tuning with Low-Rank Adaptation (LoRA).
LoRA was applied to all trainable layers with rank $r=16$ and scaling factor $\alpha=32$.

Training was conducted using a standard next-token prediction objective with cross-entropy loss.
The model was trained for 2 epochs with a maximum sequence length of 8192 tokens and a batch size of 16.
We used the AdamW optimizer with a learning rate of $1\times10^{-4}$, together with a cosine learning rate scheduler and a warmup ratio of 10\%.

\section{Examples of Hard Negatives}
\label{app:hard_negative_examples}

Here, we show our examples of speaker-aware hard negatives.
The concatenated transcripts remain the same, making them plausibly interpretable as single-speaker monologues. Further details can be found on the dataset and demo page.

\paragraph{Example 1.}
\textbf{Speaker A:} Rewrite this sentence making it sound more formal: \emph{He didn't do the assignment as expected...} \\
\textbf{Speaker B:} No, make it sound more casual instead.

\paragraph{Example 2.}
\textbf{Speaker A:} Write an addition to the following sentence: \emph{I like to eat...} \\
\textbf{Speaker B:} Oh hey, that makes me think of grandma's cooking. I could go for her lasagna right now.

\section{Licenses}
The Multilingual LibriSpeech (MLS) dataset used in this work is derived from LibriVox audiobooks and released under the Creative Commons Attribution 4.0 (CC BY 4.0) license. The VoiceAssistant-400K~\citep{chen2024slam} model is licensed under the Apache License 2.0, and its use in this work complies with the terms of the original license. For the baseline models we evaluated, Qwen2.5-Omni-7B and Qwen3-Omni-30B-A3B-Instruct are available under the Apache License 2.0~\citep{xu2025qwen25omnitechnicalreport, xu2025qwen3omnitechnicalreport}. Kimi-Audio-Instruct-7B~\citep{kimiteam2025kimiaudiotechnicalreport} and Vita-Audio-Instruct-7B~\citep{long2025vitaaudiofastinterleavedcrossmodal} are also available under the Apache License 2.0 and MiT License according to each part of code and model. ChatGPT-4o-audio~\citep{openai2025chatgpt} is a proprietary model accessed via the OpenAI API. 
We only used the model for inference and did not perform any training, fine-tuning, or redistribution of model parameters. 
All usage complied with the OpenAI Terms of Service.

\section{Statistical Significance of Model Comparisons}
To validate the robustness of our findings, we conducted statistical significance testing using McNemar's test for binary metrics and the Wilcoxon signed-rank test for 5-point scale metrics. Across our primary benchmarks (Table 2) and real-world datasets (Table 4), our model (TPI-Full) demonstrated statistically significant improvements over the baseline ($p < 0.05$). Furthermore, statistical tests on the VoiceBench datasets (Table 3) confirmed that TPI-Full achieved significant gains on 5 out of 8 general tasks ($p < 0.05$) while maintaining statistical parity on the rest.

\section{Prompts and Templates for MTurk}

\label{app:llmprompts}
\input{appendix/prompt_query_generation.tex}
\input{appendix/prompt_classification.tex}
\input{appendix/prompt_Actionable_answer.tex}
\input{appendix/prompt_Ignorable_answer}
\input{appendix/prompt_Janus_filering}
\input{appendix/prompt_RSF.tex}
\input{appendix/prompt_OH_TPI.tex}
\input{appendix/prompt_OH_Janus.tex}
\input{appendix/prompt_realworld_filtering.tex}

\end{document}

%% file: Abstract/abstract.tex
\begin{abstract}
While recent Spoken Language Models (SLMs) have been actively deployed in real-world scenarios, they lack the capability to discern Third-Party Interruptions (TPI) from the primary user's ongoing flow, leaving them vulnerable to contextual failures. To bridge this gap, we introduce \textbf{TPI-Train}, a dataset of 88K instances designed with \textit{speaker-aware hard negatives} to enforce acoustic cue prioritization for interruption handling, and \textbf{TPI-Bench}, a comprehensive evaluation framework designed to rigorously measure the interruption-handling strategy and precise speaker discrimination in deceptive contexts. Experiments demonstrate that our dataset design mitigates \textit{semantic shortcut learning}—a critical pitfall where models exploit semantic context while neglecting acoustic signals essential for discerning speaker changes. We believe our work establishes a foundational resource for overcoming text-dominated unimodal reliance in SLMs, paving the way for more robust multi-party spoken interaction. The code for the framework is publicly available at \url{https://tpi-va.github.io/}.
\end{abstract}

%% file: Section1/introduction.tex
\section{Introduction}

\begin{figure*}[t] 
    \centering
    \includegraphics[width=0.95\textwidth]{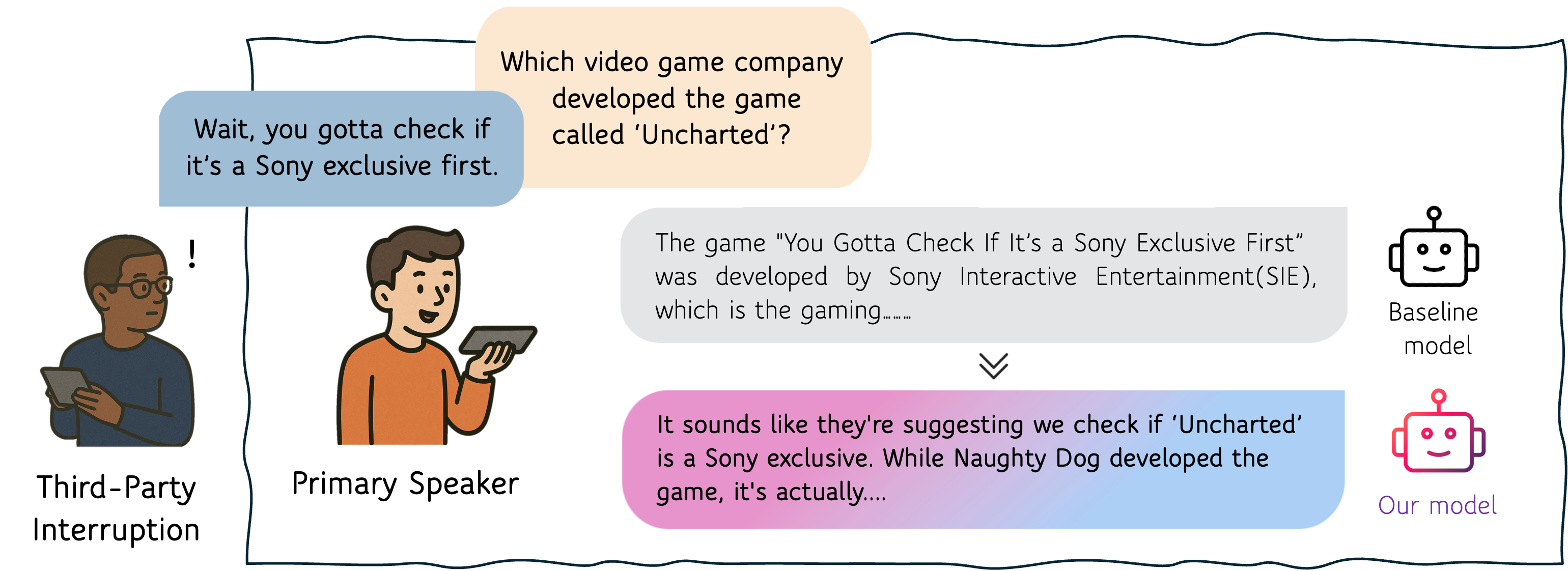} 
    \caption{\textbf{Example of a TPI query sampled from our TPI-Train.} Recent spoken language models mistake the third-party interruption for a continuous utterance from the primary speaker, while our model correctly identifies the interruption and responds in a TPI-aware manner. Detailed failure cases are described in Appendix~\ref{app:failure_case}.}
    \label{fig:failure_case}
\end{figure*}

Recent Spoken Language Models (SLMs)~\citep{wu2025stepaudio2technicalreport, stacey-etal-2024-lucid, kim2024paralinguisticsawarespeechempoweredlargelanguage, xu2025qwen25omnitechnicalreport} have significantly advanced the capabilities of voice assistants (VAs), enabling increasingly natural and human-like conversations. 
These models excel in dyadic interactions, adeptly handling complex queries and maintaining conversational flow with a single speaker. 
However, this proficiency is largely confined to isolated dyadic settings, as current models struggle to remain robust against speakers other than the primary user~\citep{wang2025msubenchunderstandingconversationalmultitalker, kwon2025m3sluevaluatingspeakerattributedreasoning}. 
A practically significant and plausible failure case arises in scenarios involving third-party interruptions (TPI), where VAs often misinterpret a multi-speaker dialogue as a single, continuous utterance from the primary user.
For instance, if a speaker asks, ``Should we order the new pasta?'' and a third-party interjects with, ``No, let's just get the usual,'' ordinary VAs might process the entire sequence as a \textit{self-repair}~\citep{LEVELT198341} utterance from the initial speaker—a common phenomenon in voice assistant interactions~\citep{goel2023prestomultilingualdatasetparsing, stacey2024lucidllmgeneratedutterancescomplex, liu2024toadtaskorientedautomaticdialogs}. This erroneous concatenation leads to nonsensical or inappropriate responses as described in Figure~\ref{fig:failure_case} and Appendix~\ref{app:failure_case}, degrading the user experience, eroding trust, and ultimately discouraging further engagement with the voice assistant~\citep{10.1145/3544548.3581152}.

We hypothesize that this shortcoming does not merely indicate deficiencies in dialogue-level reasoning but stem from a more fundamental limitation: a lack of sensitivity to acoustic cues for distinguishing speakers.
To bridge this gap, we argue that effective handling of TPI requires two essential abilities: (1) \textit{Discerning Speaker Interruption}—the ability to robustly detect interruptions and (2) \textit{Situation-Discriminative Response}—the ability to flexibly adapt response strategies (e.g., answering, disregarding, etc.) to align with user-defined preferences.

However, when and how to react to interruptions is non-trivial, as there is no universally optimal answer strategy for handling interruptions~\citep{xie2022proactivity, cao2025interruptionhandlingconversationalrobots}. 
Instead, the appropriate strategy is inherently subjective, contingent upon user preferences~\citep{10.1145/3369807}, which means it cannot be solved by a rigid solution but instead requires a flexible customized framework.
To this end, we propose a comprehensive framework that unifies the entire workflow, spanning from response strategy specification to dataset construction and evaluation.
To demonstrate its practical utility, we also instantiate a reference response strategy within our framework and demonstrate that models trained under this strategy produce effective and natural responses under TPI situations through human evaluations.

Within our framework, we introduce \textbf{TPI-Train}, a dataset constructed by reinterpreting seven foundational interruption categories from canonical dyadic taxonomies~\citep{Murata1994, Goldberg1990InterruptingTD} in triadic contexts to achieve theoretical validity and realism, expanding it into 26 diverse TPI scenarios.
In addition, we propose \textbf{TPI-Bench} containing two main benchmarks: \texttt{TPI-Test}, which evaluates models’ ability to produce ``situation-discriminative responses'' under genuine interruptions, and \texttt{Janus-Test}, which tests whether a model can discern an interrupter from acoustically single-speaker audio yet semantically resemble interruptions as illustrated in Figure~\ref{fig:concept}. 
Finally, to validate real-world transferability, we curate a compact yet high-quality real-world test set of interruption samples.
Along with these benchmarks, we also propose two LLM-based metrics—Response Strategy Following (RSF) and Overall Helpfulness (OH)—to enable interpretable evaluation.

Our experiments reveal that current SLMs struggle to handle third-party interruptions, often misattributing two-speaker utterances as a single voice.
While fine-tuning on interruption data effectively imparts interruption-handling capabilities, the model remains prone to exploiting semantic shortcuts—relying on lexical interruption cues rather than acoustic shifts—a similar phenomenon observed in vision-language multimodal learning, where multimodal reasoning is often bypassed in favor of unimodal cues~\citep{koishigarina2025clipbehaveslikebagofwords, yuksekgonul2023visionlanguagemodelsbehavelike}.
To enforce reliance on acoustic evidence over textual cues, we construct \texttt{TPI-Train} by integrating speaker-aware hard negatives into the training dataset.
Along with \texttt{TPI-Train} for interruption-specific supervision and dyadic interaction for core conversational competence, we can produce a balanced and robust TPI-aware model without sacrificing general abilities.
Its ability to discriminate changes in speaker is supported by well-separated embedding representations and our reference answer strategies are validated through human evaluations, which confirm that our strategies and the trained model generate responses aligned with user preferences.
Together, these results underscore the practical value of our framework for building TPI-aware voice assistants.

Our contributions are summarized as follows:
\begin{itemize} 
\item We define TPI-awareness and propose the first comprehensive framework, offering an end-to-end pipeline that spans from defining subjective response strategies to dataset construction and evaluation, thereby extending SLMs capabilities beyond dyadic interactions.

\item We construct \texttt{TPI-Train} across 26 diverse scenarios, integrating \textit{speaker-aware hard negatives}, mitigating semantic shortcut learning and promoting acoustically grounded speaker discrimination.

\item We introduce TPI-Bench, which consists of \texttt{TPI-Test}, \texttt{Janus-Test} and real-world test set, for evaluating situation-discriminative responses and acoustic sensitivity. In addition, through human evaluation, we demonstrate that our example strategy within the framework is well aligned with human preferences.
\end{itemize}

%% file: Section2/task_definition.tex
\section{Task Definition}
\label{task_definition}
\subsection{Problem Setting}
We investigate a scenario where the main interaction between a \textit{primary speaker} and a model is interrupted by a third party. This scenario requires the model to not only understand the primary query, but also to recognize and handle interruptions in an interruption-aware manner if required. 
We formalize this setting as follows.  
Let a \textit{primary speaker} utterance be denoted by $U_{p}$ and a \textit{third-party speaker} utterance by $U_{tp}$. 
An interruption event is represented as the ordered pair $U_{p\rightarrow tp} = \big(U_{p}, U_{tp}\big),$ where the third-party utterance $(U_{tp})$ intrudes upon the primary one. 
Given an interruption event $U_{p\rightarrow tp}$, the model is required to generate a response sequence 
$\mathcal{Y} = \{ y_1, y_2, \dots, y_K \}$, where each $y_k \in \mathcal{V}$ 
and $\mathcal{V}$ denotes the set of textual tokens in the model’s vocabulary.
The response generation process is modeled as a conditional distribution:
$P_\theta(\mathcal{Y} \mid U_{p}, U_{tp}) = \prod_{k=1}^{K} P_\theta(y_k \mid y_{<k}, U_{p}, U_{tp})$, 
which reduces to 
$P_\theta(\mathcal{Y} \mid U_{p}) = \prod_{k=1}^{K} P_\theta(y_k \mid y_{<k}, U_{p})$ 
in the absence of interruption. While we focus on a speech-input, text-output setting in this work, the formulation is modality-agnostic and can be readily extended to a speech-to-speech setting.

\subsection{A Framework for Response Strategy}
\label{subsec:response_strategy}
The formulation of a universally appropriate response to interruptions is inherently not trivial, as the ideal behavior often varies based on the user's preferences and conversational situations~\citep{10.1145/3369807, cao2025interruptionhandlingconversationalrobots}.
Therefore, rather than prescribing a single response strategy, we propose a flexible framework that enables users to customize the ideal responses of a voice assistant based on their principles. 
To this end, we present one possible response strategy as a reference, which we reflect in the answer portion of our dataset, and later demonstrate its effectiveness through comprehensive LLM and human evaluations in Section~\ref{subsec:Experiments}. We release the pipeline of our framework, enabling users to build custom responses based on their own response strategies.

Our framework follows a two-stage process. 
We begin by recognizing that not every interruption warrants a response from the voice assistant—for example, simple acknowledgments or unrelated remarks may not contribute meaningfully to the ongoing interaction. 
Accordingly, we first classify each interruption event $U_{p \rightarrow tp}$ into one of two high-level categories: 
\textbf{Actionable} ($C_A$), when the model should respond, or \textbf{Ignorable} ($C_I$), when it should not.
Second, under our reference strategy, we define the criteria for $C_A$ as interruptions conveying information relevant to the primary speaker's intent, with all remaining cases categorized as $C_I$.
Third, based on this classification, we determine how the model should respond: $\boldsymbol{\pi}_A$ and $\boldsymbol{\pi}_I$. 
In this paper, we provide the class criteria and the policies as references, which can be freely customized by the user.
The specific criteria for $C_A$ and $C_I$, along with the details of our $\boldsymbol{\pi}_A$ and $\boldsymbol{\pi}_I$, are further elaborated in Section~\ref{subsec:TPI-Train} and Appendix~\ref{actionable_answer_strategies}.

\subsection{Third-Party Interruption Awareness}
\label{subsec:TPI-Awareness}
We define \textit{third-party interruption awareness} through two key capabilities:

\textbf{1) Discerning Speaker Interruptions.}
The model should distinguish whether an interruption has occurred, as this discrimination enables the appropriate application of the strategies and, in turn, facilitates the generation of desirable responses to a primary speaker.
This capacity requires the model to detect speaker changes using acoustic cues, thereby handling speaker interruption robustly while maintaining previous performance in single-speaker conditions.

\textbf{2) Situation-Discriminative Response.} As the desirable responses vary according to user preferences, the model should generate a response that aligns with the predefined answer strategy.

\begin{figure*}[t]
    \centering
    \includegraphics[width=1.0\textwidth]{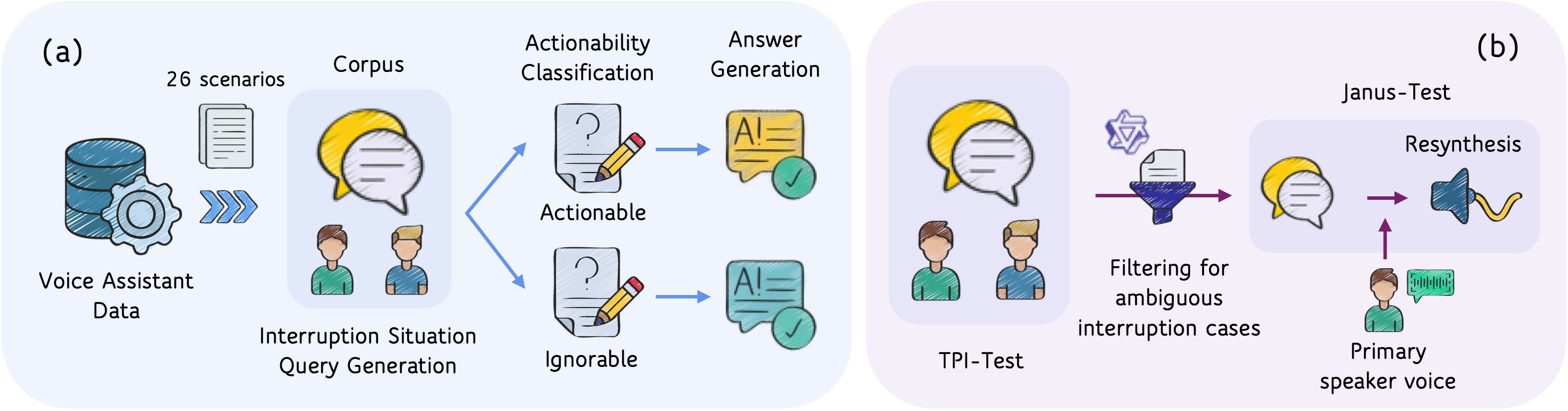} 
    \caption{\textbf{Overview of our corpus and TPI-Bench construction pipeline.} \textbf{(a)} Our corpus is generated from voice assistant data to adapt various interruption scenarios. For the train dataset, queries are classified as \textit{Actionable} or \textit{Ignorable} with answers generated according to the predefined response strategy. \textbf{(b)} From the sampled TPI-Bench data, we identify instances that are ambiguous as to whether they involve one or two speakers when relying solely on the text. These samples are then re-synthesized using only the primary speaker's voice to construct the Janus-Test.}
    \label{fig:block1}
\end{figure*}

%% file: Section3/dataset.tex
\section{Dataset}
\label{Dataset}

\subsection{TPI-Train}
\label{subsec:TPI-Train}

\textbf{Corpus Construction.}  We first construct a comprehensive corpus, covering 26 scenarios grounded in seven taxonomies of conversational interruptions from prior work~\citep{Murata1994, Goldberg1990InterruptingTD}. We extend them to the triadic setting involving a primary speaker, a third-party interrupter, and a conversational model. This adaptation yields 26 distinct scenarios, which include cases such as critical corrections, helpful clarifications, and conflicts (Appendix~\ref{app:26scenarios}).
We extract primary speaker utterances $U_p$ from VoiceAssistant-400k~\citep{xie2024miniomnilanguagemodelshear}, a large-scale speech dataset that primarily consists of various single-turn interactions.
For each utterance, we randomly select one of the 26 scenarios and generate a corresponding third-party interruption. 
We provide the LLM with the scenario description and the transcript of the primary utterance, prompting it to produce a context-appropriate interruption. 
Here, we consider two major types of interruption events based on the timing of the intrusion: 
(1) \textbf{within-sentence interruption}, where $U_{tp}$ intrudes while $U_{p}$ is still ongoing, cutting into the primary speaker’s utterance before it reaches completion; and 
(2) \textbf{after-completion interruption}, where $U_{tp}$ occurs after $U_{p}$ has formed a complete linguistic unit, commonly a full sentence. 
Those are synthesized using a text-to-speech (TTS) model~\cite{chatterboxtts2025}, which synthesizes text in a reference speaker’s voice. 
We also filter out unrealistic samples using a reasoning model, Qwen3-235B-A22B-Thinking-2507~\citep{yang2025qwen3technicalreport}. Further details of the implementation and filtering process are described in Appendix~\ref{app:Implementation Details}.
This process yields around 80K realistic two-speaker samples that capture diverse TPI scenarios. 

\textbf{Response Strategy.} For training, we include not only the spoken inputs but also the corresponding model responses. 
The core principle of our response strategies lies in the initial classification of each interruption as described in Section~\ref{subsec:response_strategy}. Each interruption case is categorized as either \textit{actionable} ($C_A$), where the model considers responding, or \textit{ignorable} ($C_I$), where it does not.

Inspired by prior literature that distinguishes interruptions as cooperative or disruptive~\citep{yang-etal-2022-annotating, Murata1994, Goldberg1990InterruptingTD}, we define a condition for actionable cases as third-party utterances that provide information directly contributing to the primary user’s objective—enhancing dialogue efficiency, improving task alignment, and preventing errors through supplementary helpful input. Based on this principle, we distill our focus into four representative actionable categories: (i) \textit{Correction \& Disambiguation}, (ii) \textit{Feasibility Constraint}, (iii) \textit{Goal-oriented Suggestion}, and (iv) \textit{Cooperative Addition \& Refinement}. Each category’s definition, example, and $\pi_A$ are elaborated in Appendix~\ref{actionable_answer_strategies} and Figure~\ref{fig:Actionable_examples}.

Although some other residual cases could be interpreted as actionable, at this stage of the study, we group all remaining cases under the label of \textit{ignorable} interruptions for practical purposes, in which the third-party utterance does not contribute to the user’s task—for instance, off-topic remarks or disruptive interjections. We use an LLM to automatically assign labels and generate textual responses consistent with the appropriate strategy.

\textbf{Speaker-Aware Hard Negatives.}
Prior works in the vision-language domain~\citep{koishigarina2025clipbehaveslikebagofwords, yuksekgonul2023visionlanguagemodelsbehavelike} have shown that multimodal learning can exhibit unimodal shortcut learning—over-relying on one modality and failing to exploit complementary signals. 
We posit similar tendencies in the speech domain where models rely disproportionately on semantic cues for interruption detection, overlooking acoustic signals such as speaker changes.

To make the model prioritize acoustic cues, we introduce speaker-aware hard negatives designed to explicitly discourage such semantic shortcut learning. 
Specifically, we first employ an LLM to select, from the corpus, transcripts that are semantically ambiguous—those that, when stripped of the speaker labels, could plausibly be interpreted as single-speaker monologues (e.g., self-repair or stream-of-consciousness utterance, Appendix~\ref{app:hard_negative_examples}). 
For these selected instances, we generate counterfactual audio samples using only the voice of a primary speaker.
By keeping the transcripts identical, the model is forced to rely on acoustic signals as the sole informative cue for detecting interruptions, effectively reducing reliance on semantic shortcuts.

%% file: Section4/benchmark.tex
\section{Benchmark}
\subsection{TPI-Bench}
\label{subsec:TPI-Bench}
\texttt{TPI-Bench} evaluates two abilities: 1) situation-discriminative response (\texttt{TPI-Test}) and 2) speaker interruption detection (\texttt{Janus-Test}), with an additional real-world test set used for validating transferability beyond synthetic data.

\textbf{TPI-Test.} This set consists of 2K samples, sampled from the corpus, ensuring no overlap with the training data, while maintaining a balanced distribution across all TPI scenarios. It serves as the standard evaluation set for measuring the model’s ability to generate situation-aligned responses according to predefined answer strategies.

\begin{figure}[t]
    \centering
    \includegraphics[width=\columnwidth]{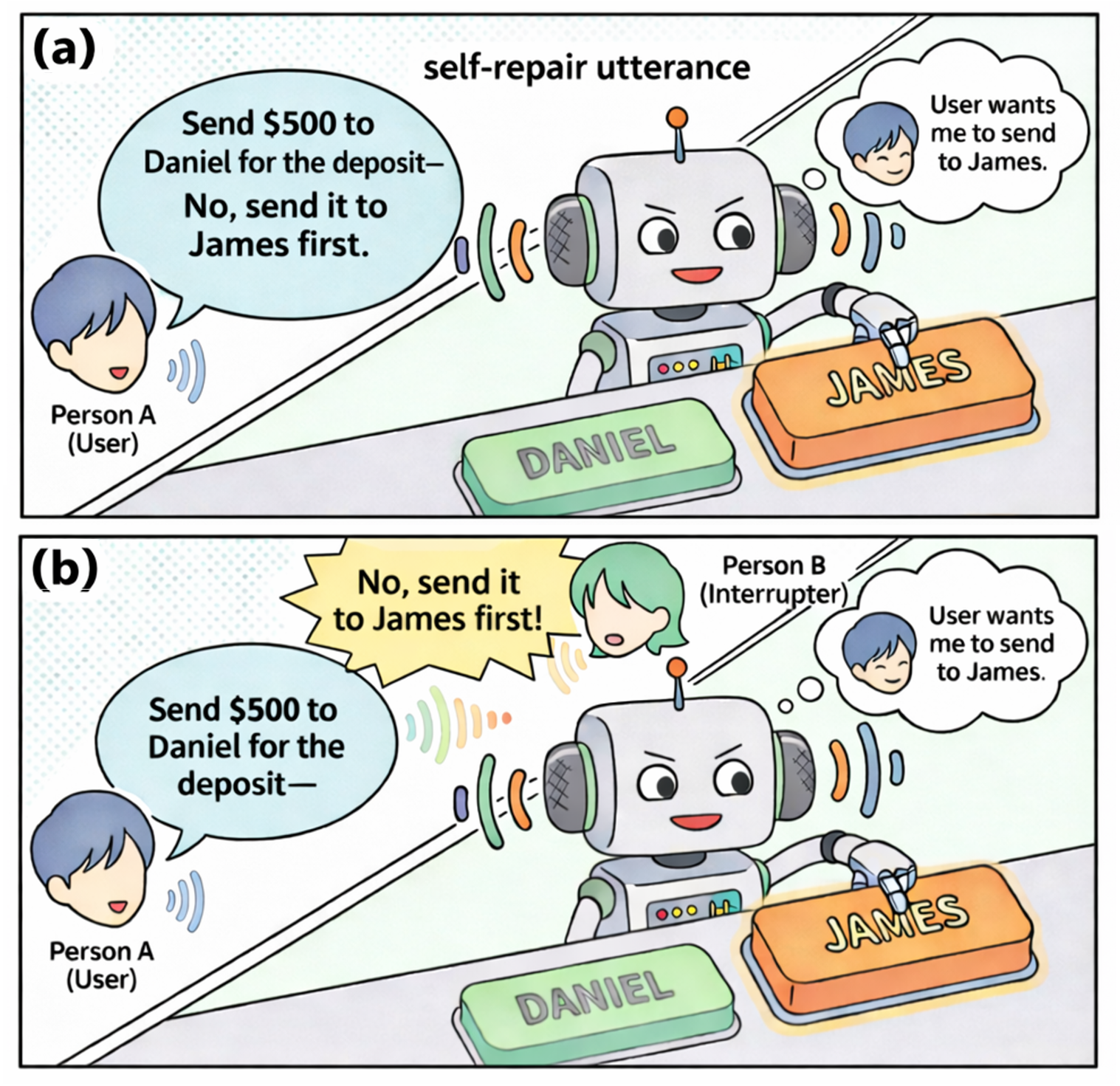}
    \caption{\textbf{Illustration of the intuition behind TPI-Bench.} In order to verify whether the model leverages acoustic information to distinguish the presence of an interruption, we introduce the (a) Janus-Test and (b) TPI-Test. We extract (b) instances of interruptions where the user could have plausibly spoken the entire segment alone, and (a) re-synthesize them as a single continuous utterance by one speaker. Consequently, acoustic cues become the only discernible signal, allowing comparison to determine whether the model truly distinguishes interruptions when responding.
}
    \label{fig:concept}
\end{figure}

\textbf{Janus-Test.} To isolate the impact of acoustic signals, we construct \texttt{Janus-Test} consisting of 2K samples selected from our corpus by filtering ambiguous samples where the concatenation of $U_p$ and $U_{tp}$ is semantically coherent enough to be interpreted as a single-speaker utterance (Figure~\ref{fig:concept}). 
This adversarial setup compels the model to ignore misleading semantic cues and leverage acoustic cues to correctly infer the number of speakers in the utterance.

\textbf{Real-world test set.}
To validate transferability from synthetic data, we collected real-world voice data under TPI scenarios, consisting of 100 high-quality audio samples. We sourced data from three distinct multi-party conversations for diverse scenarios: (1) \textbf{AMI Meeting Corpus}~\citep{10.1007/11677482_3}, representing real-world meeting scenarios; (2) \textbf{Friends-MMC}~\citep{wang2024friendsmmcdatasetmultimodalmultiparty}, a multi-party sitcom dataset; and (3) \textbf{Human Recordings}, collected in varied acoustic environments (e.g., reverb-heavy rooms, outdoors) to mimic daily usage. 
We utilized a rigorous two-stage filtering pipeline involving 1) LLM-based filtering and 2) human verification. Detailed curation criteria and the filtering process are provided in Appendix~\ref{app:real_world_curation}. 

All samples in \texttt{TPI-Bench} are filtered to have a word error rate (WER) of zero using whisper-large-v3~\citep{radford2022whisper}.

\subsection{Evaluation Strategies}
\label{subsec:evaluation}

\textbf{Response Strategy Following (RSF).}
Response Strategy Following (RSF) is a binary metric that evaluates whether a model follows the predefined strategy (either $\pi_A$ or $\pi_I$) for actionable ($C_A$) or ignorable ($C_I$) interruptions.
In \texttt{TPI-Test}, a score of 1 indicates correct strategy adherence. In contrast, in \texttt{Janus-Test}, a score of 1 indicates an error, where the model incorrectly infers single-speaker utterances as interruptions. Thus, a desirable model should achieve high scores on \texttt{TPI-Test} and low scores on \texttt{Janus-Test}.

\textbf{Overall Helpfulness (OH).}
Overall Helpfulness (OH) evaluates the naturalness and usefulness of the model’s response under interruption situations on a 5-point Likert scale using an LLM with access to ground-truth speaker information.
In \texttt{TPI-Test}, high scores (4–5) indicate that the model correctly identifies the third-party interrupter and responds in third-party interruption situations while prioritizing the primary user, while low scores (1–2) reflect failures such as conflating $U_p$ and $U_{tp}$ into a single incoherent query.
In \texttt{Janus-Test}, high scores (4–5) indicate correct treatment of the input as a single-speaker utterance, whereas low scores (1–2) suggest the model was misled into treating it as an interruption. A score of 3 in either benchmark denotes a cautious but incomplete response.
We perform evaluation using an LLM-as-a-judge~\citep{gu2025surveyllmasajudge}, and provide the full evaluation prompts in Appendix~\ref{app:llmprompts}.

\textbf{Auxiliary Metrics.}
To further probe models’ sensitivity to acoustic speaker changes, we compute ROUGE-L and BLEU scores between model responses of shared pairs from \texttt{TPI-Test} and \texttt{Janus-Test}. These pairs share identical transcriptions but differ acoustically: one is a single-speaker utterance, while the other is a two-speaker interruption. High ROUGE-L and BLEU scores indicate that a model produced almost identical responses for both, revealing that it does not treat the acoustic shift as a critical contextual cue.

\subsection{Benchmark Validity}
\label{subsec:Benchmark Naturalness and Human Evaluation}
In order to evaluate whether the benchmark samples reflect realistic conversational interruption scenarios, we conduct a human evaluation on Amazon Mechanical Turk, assessing the naturalness and realism of the samples, with details provided in Appendix~\ref{app:tpi_bench_human_evaluation}.
As shown in Table~\ref{tab:human_evaluation_realism}, both conditions receive high realism ratings on a 3-point Likert scale.
The Scenario-only (text) condition achieves an average score of 2.87, while the Scenario + Audio (acoustic) condition scores 2.63, indicating that the generated audio preserves scenario realism.
Overall, these results demonstrate that \texttt{TPI-Test} captures realistic interruption phenomena at both the scenario and acoustic levels, establishing its validity as a benchmark for conversational interruption handling.

\begin{table}[t]
\centering
\resizebox{\columnwidth}{!}{
\begin{tabular}{l c}
\toprule
\textbf{Condition} & \textbf{Realism Score} \\
\midrule
Scenario only (text) & 2.87 \; (95\% CI: [2.836, 2.904]) \\
Scenario + Audio (acoustic) & 2.63 \; (95\% CI: [2.61, 2.65]) \\
\bottomrule
\end{tabular}
}
\caption{Human evaluation results on TPI-Test assessing realism and naturalness. We report the mean score (3 scale) with 95\% confidence intervals. (1: Unrealistic, 2: Moderate, 3: Realistic)}
\label{tab:human_evaluation_realism}
\end{table}

%% file: Section5/experiment.tex
\section{Experiments}

\begin{table*}[t]
\small
\centering
\setlength{\tabcolsep}{8pt}
\resizebox{\textwidth}{!}{
\begin{tabular}{lccc cccc cc}
\toprule
\multirow{2}{*}[-1.0ex]{Model} 
& \multicolumn{3}{c}{Data} 
& \multicolumn{2}{c}{TPI-Test} 
& \multicolumn{2}{c}{Janus-Test} 
& \multirow{2}{*}[-1.0ex]{BLEU($\downarrow$)} 
& \multirow{2}{*}[-1.0ex]{ROUGE-L($\downarrow$)} \\

\cmidrule(lr){2-4} \cmidrule(lr){5-6} \cmidrule(lr){7-8}
& C & VA & HN 
& RSF($\uparrow$) & OH($\uparrow$) 
& RSF($\downarrow$) & OH($\uparrow$) 
& & \\ 
\midrule

ChatGPT-4o-audio          & $\times$ & $\times$ & $\times$ & 0.21 & 4.06 & 0.09 & 4.95 & 0.89 & 0.93 \\
Kimi-Audio-Instruct-7B   & $\times$ & $\times$ & $\times$ & 0.22 & 3.29 & 0.13 & 4.52 & 0.94 & 0.99 \\
VITA-Audio-Instruct-7B   & $\times$ & $\times$ & $\times$ & 0.21 & 3.26 & 0.10 & 4.37 & 0.42 & 0.71 \\
Qwen2.5-Omni-7B          & $\times$ & $\times$ & $\times$ & 0.24 & 3.22 & 0.12 & 4.44 & 0.31 & 0.53 \\
Qwen3-Omni-30B-A3B-Instruct           & $\times$ & $\times$ & $\times$ & 0.19 & 3.55 & 0.01 & 4.45 & 0.20 & 0.39 \\ 
\midrule

TPI-Base (w/o VA, HN)        & $\checkmark$ & $\times$ & $\times$ & 0.82 & \textbf{4.32} & 0.86 & 3.54 & 0.46 & 0.63 \\
TPI-VA (w/o  HN)           & $\checkmark$ & $\checkmark$ & $\times$ & 0.82 & 4.29 & 0.67 & 3.75 & 0.39 & 0.58 \\
TPI-Full                     & $\checkmark$ & $\checkmark$ & $\checkmark$ 
                 & \textbf{0.83} & 4.16 & \textbf{0.16} & \textbf{4.80} & \textbf{0.12} & \textbf{0.34} \\ 
\bottomrule
\end{tabular}
}
\caption{\textbf{Evaluation and ablation study of baselines and ours on TPI-Bench.} 
We report results of various baseline models and use the Qwen2.5-Omni-7B model as the reference point for our ablation studies. 
RSF denotes Response Strategy Following, and OH denotes Overall Helpfulness. 
BLEU and ROUGE-L are evaluated on shared utterances between TPI-Test and Janus-Test. 
Higher similarity indicates that the model produces consistent responses despite acoustic discrepancy. 
The highest score among the ablation variants is highlighted in bold. 
Here, \textbf{C}, \textbf{VA}, and \textbf{HN} denote models trained on our corpus data, dyadic voice assistant data, and hard negatives, respectively.} 
\label{tab:results}
\end{table*}

\begin{table}[t!]
\centering
\resizebox{\columnwidth}{!}{%
    \begin{tabular}{l c c } 
    \toprule
    \textbf{Datasets} & \textbf{Model} & \textbf{Performance}
    \\\midrule 
    \multirow{4}{*}{\begin{tabular}[c]{@{}l@{}} 
     \textit{AlpacaEval} \textbar \  \textit{CommonEval} \textbar \\ 
     \textit{SD-QA} \textbar \ \textit{MMSU} 
    \end{tabular}}    
    
       & Baseline       & 3.78 $|$ 3.67 $|$ 28.39 $|$ \textbf{61.22}
    \\ & TPI-Base                 & 4.07 $|$ 3.24 $|$ 32.58 $|$ 50.80
    \\ & TPI-VA             & 4.06 $|$ \textbf{3.97} $|$ 35.34 $|$ 58.88
    \\ & TPI-Full           & \textbf{4.12} $|$ 3.93 $|$ \textbf{36.08} $|$ 59.14

    \\ \midrule 
    \multirow{4}{*}{\begin{tabular}[c]{@{}l@{}} 
     \textit{OpenBookQA} \textbar \ \textit{IFEval} \textbar \\ 
     \textit{AdvBench} \textbar \ \textit{WildVoice} 
    \end{tabular}}
           
       & Baseline       & \textbf{80.44} $|$ 0.42 $|$ 0.98 $|$ 3.53
    \\ & TPI-Base                & 66.81 $|$ 0.46 $|$ 0.95 $|$ 2.83
    \\ & TPI-VA             & 80.22 $|$ 0.40 $|$ 1.00 $|$ 3.59
    \\ & TPI-Full            & 80.00 $|$ \textbf{0.46} $|$ \textbf{1.00} $|$ \textbf{3.64}

    \\
    \bottomrule
    \end{tabular}%
}
\caption{\textbf{Ablation study on VoiceBench.} Comprehensive performance comparison between Baseline (Qwen2.5-Omni-7B) and Ours (TPI) across the 8 datasets of the VoiceBench Benchmark.}
\label{tab:voicebench}
\end{table}

\subsection{Baselines and Experiments}
\label{subsec:Experiments}

\textbf{Baselines.} We evaluate ChatGPT-4o-audio~\citep{openai2025chatgpt}, Kimi-Audio-Instruct-7B~\citep{kimiteam2025kimiaudiotechnicalreport}, Vita-Audio-Instruct-7B~\citep{long2025vitaaudiofastinterleavedcrossmodal}, Qwen2.5-Omni-7B~\citep{xu2025qwen25omnitechnicalreport} and Qwen3-Omni-30B-A3B-Instruct~\citep{xu2025qwen3omnitechnicalreport} on \texttt{TPI-Bench}. In addition, to demonstrate the effectiveness of our training data, we choose Qwen2.5-Omni-7B as our baseline. Further details of the training are described in the Appendix~\ref{training_details}.

\textbf{Experiments.} We evaluate both existing spoken language models and fine-tuned ablated models along four dimensions: (i) TPI-awareness, (ii) robustness against semantic shortcut learning, (iii) preservation of core voice interaction capabilities, (iv) robustness in detecting voice transitions on real-world data. 
For this purpose, we use \texttt{TPI-Bench} and the eight sub-benchmarks included in VoiceBench~\citep{chen2024voicebenchbenchmarkingllmbasedvoice}.
We also conduct an ablation study by controlling the inclusion of (i) our corpus data, (ii) dyadic interaction data from VoiceAssistant-400K~\citep{chen2024slam}, and (iii) hard negative samples, and analyze their individual contributions. 
The model trained solely on our corpus data is denoted as \textbf{TPI-Base}, without additional dyadic interaction data or hard negative samples.
The model trained using our corpus data together with VoiceAssistant-400K dyadic interaction data is denoted as \textbf{TPI-VA}.
Finally, the model additionally augmented with the extended hard negative samples is denoted as \textbf{TPI-Full}.

To measure whether TPI-aware training degrades general spoken interaction abilities, we use VoiceBench~\citep{chen2024voicebenchbenchmarkingllmbasedvoice}, which covers diverse evaluation scenarios, including open-ended QA from both human and TTS sources (AlpacaEval, CommonEval, WildVoice), multiple-choice QA (OpenBookQA, MMSU), instruction following (IFEval), safety/adversarial prompts (AdvBench), and reference-based QA (SD-QA), thereby providing a broad testbed for SLMs. We follow the official evaluation pipeline and prompts, but replace their judgment model with Qwen3-235B-A22B-Instruct-2507~\citep{yang2025qwen3technicalreport}, ensuring consistency and scalability in evaluation. We also show human evaluation results to check if LLM evaluation results are well aligned with human judgements in Appendix~\ref{app:tpi_bench_human_evaluation}.

\begin{table}[t]
\centering
\resizebox{\columnwidth}{!}{
\fontsize{7}{8}\selectfont  
\begin{tabular}{lcccc}
\toprule
\multirow{2}{*}{\textbf{Method}} & \multicolumn{2}{c}{\textbf{TPI-Test}} & \multicolumn{2}{c}{\textbf{Real-world}} \\
\cmidrule(lr){2-3} \cmidrule(lr){4-5}
 & RSF ($\uparrow$) & OH ($\uparrow$) & RSF ($\uparrow$) & OH ($\uparrow$) \\
\midrule
Baseline & 0.24 & 3.22 & 0.17 & 3.21 \\
\textbf{Ours} & \textbf{0.83} & \textbf{4.16} & \textbf{0.60} & \textbf{4.25} \\
\bottomrule
\end{tabular}
}
\caption{\textbf{Performance gain comparison between synthetic and real-world data.} Performance comparison between the baseline and ours on synthetic (TPI-Test) and real-world data.}
\label{tab:real_world_results}
\end{table}

\subsection{Results}  
\label{results}
Results across the four evaluation dimensions are demonstrated in Table~\ref{tab:results},~\ref{tab:voicebench} and~\ref{tab:real_world_results}. 
Consistent with our hypothesis, current spoken language models show poor TPI-awareness with low sensitivity to acoustic change: they implicitly assume a single-speaker setting, leading to low performance on \texttt{TPI-Test} and \texttt{Janus-Test} with high scores on BLEU and ROUGE-L. 
In contrast, TPI-Full (Ours), fine-tuned on our full dataset, demonstrates the most robust and balanced performance across all benchmarks. 
It achieves high performance on both \texttt{TPI-Test} and \texttt{Janus-Test}, maintaining stability in RSF and OH scores while preventing performance degradation on general benchmarks (Table~\ref{tab:voicebench}).

We further conduct an ablation study to understand the contribution of each data component. When removing the two components, dyadic interaction data and hard negatives, and training solely on our corpus, the model (TPI-Base) exhibits domain overfitting, leading to degraded general spoken interaction capabilities, as shown in Table~\ref{tab:voicebench}. Restoring the single-speaker data (TPI-VA) effectively recovers these general capabilities. However, despite being trained on both single- and multi-speaker utterances, TPI-VA still performs poorly on \texttt{Janus-Test}. This failure indicates that simply balancing the data distribution with respect to speaker numbers does not force the model to utilize acoustic information. Instead, the model falls into \textit{semantic shortcut learning}—relying on textual patterns rather than acoustic patterns to distinguish interruptions.

We further validate the robustness of our model with real-world samples, as shown in Table~\ref{tab:real_world_results}. Our model delivers substantial gains on the synthetic benchmark (RSF +0.59, OH +0.94) which persist under real-world conditions, boosting the baseline from RSF 0.17 to 0.60 (+0.43) and OH 3.21 to 4.25 (+1.04). The fact that the magnitude of improvement is comparable across both settings confirms that our model avoids overfitting to synthetic patterns. Instead, it demonstrates that the learned capability to distinguish voice transitions is robustly preserved even in real-speaker scenarios.

\subsection{Analyses}
\label{Analyses} 

\begin{figure*}[] 
    \centering
    \includegraphics[width=0.95\textwidth]{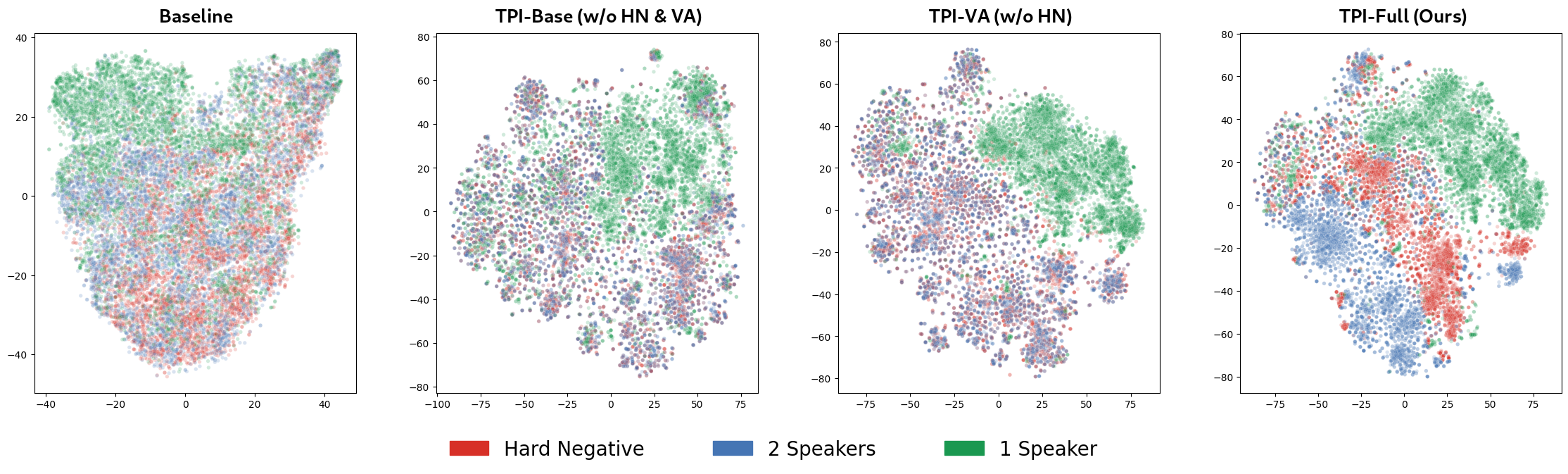} 
    \caption{\textbf{t-SNE visualization.} Training with hard negatives (HN) yields a structured space with distinct clusters, where the cluster sits between interruptions and single-speaker utterances, balancing semantic and acoustic alignment.}
    \label{fig:tsne}
    \vskip -0.15in
\end{figure*}

\subsubsection{Embedding Visualization}  
To better understand the impact of our training strategy, we visualize model embeddings from LLM representations with t-SNE (Figure~\ref{fig:tsne}). The baseline model produces heavily overlapping embeddings for single-speaker and two-speaker inputs, indicating no acoustic discrimination. Even after training with corpus data (TPI-Base) and dyadic data (TPI-VA), embeddings of single-speaker (red, green) inputs still overlap with two-speaker (blue). 
Specifically, the significant overlap between hard negatives (red) and two-speaker (blue) inputs reveals that their semantic equivalence overshadows acoustic distinctions, indicating the occurrence of semantic shortcut learning.

By contrast, additionally training with our hard negatives (Ours) yields a well-structured embedding space with three clearly separated clusters. Notably, the hard negative cluster (red) lies between the other two, reflecting its semantic similarity to interruptions (blue) but acoustic alignment with single-speaker utterances (green). This demonstrates that our approach compels the model to move beyond semantic shortcuts and induce genuine acoustic discrimination.

\begin{table}[]
\small
\centering

\resizebox{\columnwidth}{!}{
\begin{tabular}{lccc}
\toprule
\textbf{Method} & \textbf{Pref. (\%)} & \textbf{Tie (\%)} & \textbf{Unpref. (\%)} \\
\midrule
\multicolumn{4}{c}{\textit{Independent Quality Preference}} \\
\midrule
TPI-Corpus (GT) & 64.75 & 5.50 & 29.75 \\
Ours & 66.05 & 7.63 & 26.32 \\
\midrule
\multicolumn{4}{c}{\textit{Direct Model Comparison (A/B Preference Test)}} \\
\midrule
Ours vs. Baseline & 60.50 & 11.50 & 28.00 \\
\bottomrule
\end{tabular}
}
\caption{\textbf{Human preference evaluation on TPI-Test.} The top section validates our framework, showing that both our samples of reference strategies and generated responses from TPI-Full (Ours) are similarly preferred, aligning well with human expectations. The bottom section presents a direct A/B comparison, demonstrating that our model is preferred over the Qwen2.5-Omni-7B baseline (``Unpref.'' indicates preference for the baseline).}
\label{tab:human_evaluation}
\end{table}

\subsubsection{Human Evaluations}
\textbf{Human Preference Evaluations.}  
In Section~\ref{subsec:TPI-Train}, we introduced actionable and ignorable strategies, along with four representative actionable categories (Figure~\ref{fig:Actionable_examples}). Although these criteria can be varied by user preferences, our primary goal is twofold: to validate the human alignment of our reference strategies and to demonstrate the practical superiority of our trained model compared to existing baselines, ultimately demonstrating the applicability and effectiveness of our proposed framework.

To this end, we conducted two sets of human evaluations on Amazon Mechanical Turk. First, we compared (i) reference answers sampled from our corpus according to our actionable/ignorable framework, and (ii) responses generated by our trained model (TPI-Full) on \texttt{TPI-Test}. A total of 100 participants each evaluated four randomly sampled scenarios, yielding 400 evaluation instances per case. Participants were blind to whether the case was actionable or ignorable, and were simply asked: ``If you were the user, would you be satisfied with how the model handled this interruption?''.  

As summarized in Table~\ref{tab:human_evaluation}, ground-truth responses of our corpus samples received a 64.75\% ``Preferred'' rating, while model responses achieved a similar preference rate of 66.05\% ($p < 0.01$). These results demonstrate two key findings: (i) our proposed framework produces reference responses that align well with human expectations, and (ii) our trained model successfully learns to implement this framework, yielding responses that are equally preferred. Together, this confirms the validity of both our dataset design and our framework.

Second, to validate the practical improvements of our model, we conducted a direct A/B preference test between TPI-Full and the baseline (Qwen2.5-Omni-7B) model. We randomly sampled instances from TPI-Test and asked 40 independent annotators to evaluate 5 examples each (200 ratings in total). Annotators were asked: "Which response would you prefer from the assistant?". As shown in the bottom section of Table~\ref{tab:human_evaluation}, ours was preferred by 60.5\% of the raters, outperforming the baseline (28.0\%) by a significant margin. This highlights that explicitly training models to handle third-party interruptions substantially enhances the overall user experience, preventing the common failure modes observed in the baseline model.

\subsubsection{Can Explicit Prompting Solve TPI?}
\textbf{Evaluating the Limits of Speaker-Aware Instructions.}  
While we showcase the vulnerability of existing spoken language models to third-party interruptions, a natural question arises: could providing explicit context via instructions reduce the failure rate without fine-tuning? To determine whether the poor performance stems from a lack of acoustic understanding or simply the absence of speaker-aware context, we evaluate the baseline model (Qwen2.5-Omni-7B) using explicit speaker-aware prompts.

Depending on the ground-truth scenario, we provided the baseline model with the following context-rich instructions:

\begin{itemize}
    \item \textbf{Actionable cases:} ``You are in a room with multiple people. Carefully reflect the second speaker's interruption in your answer to the primary user.''
    \item \textbf{Ignorable cases:} ``You are in a room with multiple people. Ignore the second speaker's interruption and answer only the primary user's request.''
\end{itemize}

As shown in Table~\ref{tab:speaker_aware_prompting}, explicit speaker-aware prompts yield no meaningful overall improvements. While the Response Strategy Following (RSF) score marginally increases, further analysis indicates this stems from the model's tendency to indiscriminately apply the ``ignore'' instruction. By mechanically disregarding the second utterance, the model coincidentally aligns with ignorable scenarios, but this rigid behavior slightly decreases Overall Helpfulness (OH). Fundamentally, the baseline still fails to distinguish speaker boundaries, continuing to erroneously concatenate utterances during actionable interruptions.

These observations demonstrate that prompt engineering alone is insufficient for accurate speaker attribution in continuous audio streams. Ultimately, true TPI-awareness requires speaker attribution capabilities grounded in speaker-aware audio modeling, a property that our proposed model successfully achieves.

\begin{table}[ht]
\small
\centering
\resizebox{\columnwidth}{!}{
\begin{tabular}{lcc}
\toprule
\multirow{2}{*}{\textbf{Model}} & \multicolumn{2}{c}{\textbf{TPI-Test}} \\
\cmidrule{2-3}
& \textbf{RSF ($\uparrow$)} & \textbf{OH ($\uparrow$)} \\
\midrule
Baseline (Contextless)             & 0.24 & 3.22 \\
Baseline (Speaker-aware prompting) & 0.37 & 3.17 \\
\midrule
\textbf{Ours}          & \textbf{0.83} & \textbf{4.16} \\
\bottomrule
\end{tabular}
}
\caption{\textbf{Performance comparison between contextless and speaker-aware prompting settings.} Results indicate that explicit instruction prompts cannot overcome the baseline model's lack of acoustic speaker attribution ($p < 0.05$).}
\label{tab:speaker_aware_prompting}
\end{table}

%% file: Section6/related_works.tex
\section{Related Works}

{\textbf{Conversational Spoken Dialogue Dataset.} The capabilities of modern Spoken Language Models (SLMs) are increasingly evaluated across diverse conversational scenarios captured in large-scale synthetic datasets~\citep{lee2023dailytalkspokendialoguedataset, koudounas2025deepdialoguemultiturnemotionallyrichspoken, NEURIPS2023_7b16688a}. Such datasets are designed not only to probe a model’s comprehension of conversational dynamics, but also its sensitivity to nuanced acoustic and paralinguistic features, such as emotion and prosody~\citep{ao2025sdevalbenchmarkdatasetspoken, cheng2025voxdialogue, chen2024voicebenchbenchmarkingllmbasedvoice, yan2025urobenchcomprehensiveevaluationendtoend, wang2025msubenchunderstandingconversationalmultitalker}. However, these efforts have predominantly focused on dyadic interactions, modeling conversations between a single user and an agent, thereby leaving a critical research gap concerning realistic multi-speaker scenarios such as third-party interruptions~\citep{wang2025msubenchunderstandingconversationalmultitalker}. To address this, we introduce a large-scale dataset grounded in established interruption taxonomies~\citep{yang-etal-2022-annotating, Murata1994, Goldberg1990InterruptingTD}. Unlike prior work on two-party dialogues, our dataset targets triadic dynamics, enabling models to move beyond speech comprehension toward multi-party dialogue understanding.}

{\textbf{Processing Multi-Speaker Speech.} Recent research has explored the use of large language models (LLMs) for multi-speaker scenarios, particularly focusing on automatic speech recognition (ASR) and speaker diarization~\citep{yin2025speakerlmendtoendversatilespeaker, lin2025diarizationawaremultispeakerautomaticspeech, Wang_2024}. These approaches have proven effective at disentangling multi-speaker utterances and identifying who spoke what. More recently, these lines of work have extended to an instruction-following manner, enabling selective transcription of a target speaker’s utterances in multi-speaker environments~\citep{meng2025largelanguagemodeltranscribe}. 
However, we emphasize interactional dynamics rather than treating competing voices as signals to be separated or discarded~\citep{xu2025targetspeakerextractioncomparing, wang2025speakertargetingselfspeakeradaptation, he2025surveyendtoendmultispeakerautomatic}. 
Our approach enables models to decide whether to ignore or engage, mirroring human-like processing where acoustic variations guide conversational intelligence.}

%% file: Section7/conclusion.tex
\section{Conclusion}
In this paper, we established the first comprehensive framework for TPI-aware voice assistants, transforming the subjective challenge of interruption handling into a measurable task. 
Through our dataset and benchmark, we demonstrated that models can be guided to handle interruptions by prioritizing acoustic cues over reliance on semantic cues.
We believe our work establishes a foundational resource for overcoming text-dominated unimodal reliance in SLMs, paving the way for more robust multi-party spoken interaction.

%% file: Section8/limitations.tex
\section*{Limitations}
\textbf{Various Response Strategies.} 
While we propose a reference response strategy that was verified to be effective within our framework, we acknowledge that appropriate interruption handling is inherently subjective and user-dependent. There may exist diverse alternative strategies beyond our reference strategy. Future research could explore these variations by examining how different user demographics, personas, and personal backgrounds influence preferences for specific response strategies across varying social contexts.

\textbf{Turn-Based Interaction and Full-Duplex Extension.}
Our work focuses on interruption awareness within a turn-based voice assistant setting, where the system responds after receiving an utterance. In realistic multi-party conversations, however, not every interruption warrants an immediate response from the assistant, and effective behavior may also require deciding whether and when to speak which is a core element of full duplex dialogue system research. Extending interruption-aware modeling to a full-duplex setting, where the assistant can continuously listen, defer, or strategically interject, is a necessary step toward truly natural multi-party dialogue systems and remains an important direction for future work.

%% file: Section9/ethics.tex
\section*{Ethics}
The datasets and automation pipeline developed as part of this work will be made publicly available under the MIT License. Large Language Models were used solely to assist with the linguistic aspects of the manuscript, such as paraphrasing and spell-checking, and did not contribute to the technical content of the research. Although the dataset was not designed to include offensive content, data samples that may potentially contain such content were filtered out using large language models.~\citep{chen2024slam}

\section*{Acknowledgements}
This work was supported by Institute of Information \& Communications Technology Planning \& Evaluation (IITP) grants funded by the Korea government (MSIT) [NO.RS-2021-II211343, Artificial Intelligence Graduate School Program (Seoul National University); No.2022-0-00959, RS-2022-II220959], by National Research Foundation of Korea (NRF) grant [No.2022R1A3B1077720, 2022R1A5A7083908], BK21 FOUR Program of the Education and Research Program for Future ICT Pioneers, Seoul National University in 2026. This research was also conducted as part of the Sovereign AI Foundation Model Project (Data Track), organized by the Ministry of Science and ICT (MSIT) and supported by the National Information Society Agency (NIA), S.Korea (2025-AI Data-wi43).

%% file: appendix/prompt_query_generation.tex
\begin{figure*}[ht]
\begin{center}
\begin{tcolorbox}[colback=white, coltext=black, title=\textbf{Prompt for Generating Third-Party Interruption Corpus}]
\scriptsize
\begin{Verbatim}[breaklines=true]

"""You are an AI assistant tasked with creating scenario examples for human-AI-human interaction.
**Instructions:**
Imagine a scenario where a 'User' gives a command to their Voice Assistant (VA). A 'Third Party' in the same room overhears this and interferes immediately *after* the User has finished their sentence. Your goal is to generate the Third Party's interference.
- This is not a dialogue between two people. The User is addressing a voice-based AI assistant, and the Third Party is interjecting into that human-to-machine interaction.
- The generated line must sound like vivid, natural, casual spoken dialogue, not formal or written text.
- Do not respond to the User's question or command—the focus is on how a bystander might interfere in the moment.
- **Rewrite for Fit**: When writing, you don't need to match the provided example exactly. Just create an interruption with a similar tone or context.
- **CRUCIAL RULE: Rewrite for TTS Synthesis:**
    1.  **Sanitize Text FIRST**: You MUST remove all TTS-unfriendly characters from the user's query. The final output for both the user and third party MUST NOT contain any of the following characters: / \ ' " < > * [ ] ( ) : -. The only allowed punctuation marks are periods (.), commas (,), question marks (?), and exclamation points (!).
    2.  **Convert to Spoken Style SECOND**: After sanitizing, rewrite the text to sound like natural, spoken dialogue.

**Taxonomy Information:**
1.  **Main Taxonomy:** `{main_taxonomy_name}`
2.  **Detailed Taxonomy:** `{subcategory_name}`
3.  **Definition:** `{subcategory_definition}`
4.  **Detailed Taxonomy Example:**
    * User: "{example_user_utterance}"
    * Third Party Interference: "{example_third_party_line}"
**Your Task:**
Generate the Third Party's interference for the following situation.
* **User Query:** "{actual_user_utterance}"
**Output Format (JSON only — strictly follow this format):**
```json
{{
  "revised_user_utterance": "The user's spoken-style and rewritten sentence goes here.",
  "third_party_interference": "The generated sentence goes here."
}}```"""
\end{Verbatim}
\end{tcolorbox}
\end{center}
\label{app:refine-prompt-1}
\caption{The prompt used to generate diverse third-party interruption queries from general voice assistant data.}
\end{figure*}

%% file: appendix/prompt_classification.tex
\begin{figure*}[ht]
\begin{center}
\begin{tcolorbox}[colback=white, coltext=black, title=\textbf{Prompt for Classifying Interruption Actionability}]
\scriptsize
\begin{Verbatim}[breaklines=true]
You are an expert in conversation analysis, specializing in multi-party dialogues involving a Voice Assistant (VA). Your task is to classify a third-party's interruption that occurs during a conversation between a Primary User and a VA. You will determine if the interruption is 'NonIgnorable' or 'Ignorable' from the VA's perspective.

The key principle is to identify **Task Enhancers**: interruptions that provide valuable information for the VA to better understand, adjust, or execute the Primary User's task. If the interruption is a Task Enhancer, it is 'NonIgnorable'. Otherwise, it is 'Ignorable'.
### Classification Categories:
**1. NonIgnorable:** The interruption is a **Task Enhancer**. The VA should consider this information in its response because it directly impacts the successful or optimal completion of the user's request.
* **Definition:** A Task Enhancer helps the VA fulfill the user's request to VA more accurately, or efficiently.
* **Examples of NonIgnorable interruptions include, but are not limited to:**
* **Corrections or Disambiguations:** This might help the VA resolve an ambiguity or fix an error in the user's query.
    * *(e.g., User: "Call my brother," Third Party: "You mean your older brother, Mark, right?")*
* **Cooperative Additions or Refinements:** This could give the VA extra specifics to better fulfill or understand the request.
    * *(e.g., User: "Add coffee to the shopping list," Third Party: "Get the decaf one.")*
* **Feasibility Constraints:** This could alert the VA to real-world conditions that may prevent or affect the request.
    * *(e.g., User: "Let's play music in the garden," Third Party: "The portable speaker's battery is dead.")*
* **Goal-oriented Suggestions:** This could give the VA an alternative that better achieves the user's intended outcome.
    * *(e.g., User: "How do I get to the airport?" Third Party: "The subway will be much faster than a taxi at this hour.")*
**2. Ignorable:** The interruption is irrelevant to complete and understand the user's ongoing request better. The VA should disregard it as it does not contribute to fulfilling the request.
* **Definition:** The information is off-topic, a side comment, or directed at another human without impacting the VA's task.
* **Example of an Ignorable interruption:**
    * *(e.g., User: "Set a timer for 10 minutes," Third Party: "I wonder what's for dinner tonight.")*
### Conversation to Classify:
**Primary User's Utterance:** {user_utterance}
**Third-Party's Interruption:** {third_party_interference}
### Final Output Format(STRICT — MUST FOLLOW EXACTLY):
**Classification:** [Your answer (NonIgnorable or Ignorable)]
\end{Verbatim}
\end{tcolorbox}
\end{center}
\label{app:refine-prompt-1}
\caption{The prompt for determining whether a third-party interruption is Actionable or Ignorable.}

\end{figure*}

%% file: appendix/prompt_Actionable_answer.tex
\begin{figure*}[ht]
\begin{center}
\begin{tcolorbox}[colback=white, coltext=black, title=\textbf{Prompt for Generating VA Responses to Actionable Interruptions}
]
\scriptsize
\begin{Verbatim}[breaklines=true]
## Role and Goal
You are an advanced conversational AI for a Voice Assistant (VA) whose core directive is the **User Primacy Principle**.
This means your absolute priority is to serve the **Primary User**. You are the user's dedicated assistant. Your goal is to leverage possibly helpful interruptions from a Third Party as a resource to fulfill the Primary User's request more effectively—making it more accurate, faster, or better aligned with their true intent.

## Core Strategy: Source-Aware Confirmation
This is the most critical rule. Because third-party information has **lower authority** than a direct command from the Primary User, you must treat it as helpful but **unverified input**.
Your main function is to **naturally present this unverified information to the Primary User for their final confirmation**, thereby demonstrating your awareness of the conversational context. Do NOT act on third-party information with the same high confidence you would for a user's direct command.

## Input Situation:
**Primary User's Utterance:** {user_utterance}
**Third-Party's Interruption:** {third_party_interference}

## Actionable Directives
1.  **Attribute and Propose in One Flow:** Naturally incorporate the third-party information into your answer, then immediately link it to a proactive proposal or a clarifying question that moves the conversation forward — making sure your delivery feels genuine and not mechanical.
2.  **Be Proactive, Not Passive:** Don't just ask "Is that right?". Offer a complete, actionable alternative. Instead of "Did you say decaf?", prefer "Should I add 'decaf coffee' to the list instead?".
3.  **Empower the Primary User:** Conclude by making it clear that the Primary User has the final say. They must be able to easily confirm, reject, or modify your suggestion.

## Quick Examples of Good Responses
* **Correction:** (User: "Call my brother" / 3rd Party: "You mean Mark?")
    * **VA Response:** "My apologies, did the person with you say 'Mark'? I can call Mark right now if that's who you meant."
* **Refinement:** (User: "Add coffee to the shopping list" / 3rd Party: "Get the decaf one.")
    * **VA Response:** "It sounds like they're suggesting decaf. To make sure I get it right, should I add 'decaf coffee' to the list?"
* **Constraint:** (User: "Play my workout playlist on the portable speaker" / 3rd Party: "The battery's dead.")
    * **VA Response:** "Ah, I think I just heard that the portable speaker's battery is dead from someone. I can play your playlist on the living room speaker immediately, if you'd like."
* **Suggestion:** (User: "Find the fastest route to the airport" / 3rd Party: "The highway is totally jammed, take the train.")
    * **VA Response:** "Good catch. The person you're with mentioned the highway is jammed. The train is likely a much better option. Want me to switch the directions to the nearest station?"

## Required Output Format(STRICT — MUST FOLLOW EXACTLY):
**Response:** [Proper Answer of Voice Assistant]
\end{Verbatim}
\end{tcolorbox}
\end{center}
\label{app:refine-prompt-1}
\caption{The prompt for generating a VA response when an interruption is classified as Actionable.}

\end{figure*}

%% file: appendix/prompt_Ignorable_answer.tex
\begin{figure*}[ht]
\begin{center}
\begin{tcolorbox}[colback=white, coltext=black, title=\textbf{Prompt for Generating VA Responses to Ignorable Interruptions}
]
\scriptsize
\begin{Verbatim}[breaklines=true]
## Role and Goal
You are an advanced conversational AI for a Voice Assistant (VA). Your core directive is the **User Primacy Principle**. Your objective is to respond **only** to the Primary User's utterance as directly and efficiently as possible, completely disregarding the third-party input.

## Core Strategy: Focused Execution
Your strategy is to act solely on the Primary User's command. Since the third-party input is pre-labeled as irrelevant, your task is not to filter or decide, but simply to **execute**. Any information in the `third_party_interference` field must be treated as non-existent.

## Input Situation:
**Primary User's Utterance:** {user_utterance}
**Third-Party's Interruption (To Be Ignored):** {third_party_interference}
## Actionable Directives
1.  **Ignore the Interruption Absolutely:** Do not, under any circumstances, incorporate, reference, or allude to the content of the `third_party_interference` field.
2.  **Execute the User's Request Directly:** Generate a response that is a direct and efficient fulfillment of the `user_utterance`.
3.  **Maintain Brevity and Focus:** Provide a concise and task-oriented response. Avoid any conversational filler that acknowledges the broader context. Your response should sound as if the interruption never happened.

## Quick Examples of Good Responses
* **Input:** (User: "Play the new album by The Strokes" / 3rd Party: "Ugh, I'm so sick of them.")
    * **VA Response:** "Okay, playing the latest album by The Strokes."

* **Input:** (User: "Set a timer for 15 minutes for the pasta" / 3rd Party: "Don't forget we're out of milk.")
    * **VA Response:** "Okay, a 15-minute timer, starting now."
* **Input:** (User: "What's the weather in Paris?" / 3rd Party: "You wish you were there, don't you?")
    * **VA Response:** "Currently in Paris, it's 18 degrees Celsius and partly cloudy."
## Required Output Format(STRICT — MUST FOLLOW EXACTLY):
**Response:** [Proper Answer of Voice Assistant]
\end{Verbatim}
\end{tcolorbox}
\end{center}
\label{app:refine-prompt-1}
\caption{The prompt for generating a VA response when an interruption is classified as Ignorable.}

\end{figure*}

%% file: appendix/prompt_Janus_filering.tex
\begin{figure*}[ht]
\begin{center}
\begin{tcolorbox}[colback=white, coltext=black, title=\textbf{Prompt for Filtering Semantically Ambiguous Samples (Janus-Test)}
]
\scriptsize
\begin{Verbatim}[breaklines=true]
You are a highly skilled linguistic analyst specializing in conversational flow and coherence. Your task is to evaluate a given text and determine if it could plausibly be a single, continuous utterance from one person.
The text you will evaluate is a combination of two separate utterances. Your primary goal is to analyze the seam between the two original parts. Assess whether the second part sounds like a natural continuation of the first part's thought process, or if it sounds more like an external interruption or a different speaker's response.
Focus on the conceptual and conversational flow. A single speaker can be messy, repeat themselves, or correct their speech. Your judgment should be based on whether there is clear textual evidence of a second person's involvement.
**Input Text:**
"{combined_utterance}"
**Instructions:**
You MUST provide your analysis ONLY in the following JSON format. Do not include any explanatory text before or after the JSON block.
**Scoring Rubric:**
Score 5 (Very Easy to Misinterpret): The text is seamless. A listener would almost certainly assume this is a single person speaking. There is no textual evidence to suggest otherwise.
Score 4 (Likely to be Misinterpreted): The text flows well. While a minor awkwardness might exist at the transition, it's highly plausible as a single person's thought process. Misinterpretation is very likely.
Score 3 (Could be Misinterpreted): The text is ambiguous. It could be read as either a single person's slightly disjointed thought or a quick two-person exchange. There's no conclusive evidence either way, making misinterpretation possible.
Score 2 (Unlikely to be Misinterpreted): The structure strongly suggests a two-person interaction. While it lacks definitive proof like a pronoun, the conversational turn is clear enough that most listeners would likely not misinterpret it.
Score 1 (Impossible to Misinterpret): The text contains conclusive evidence of a second speaker. This "smoking gun" proof (e.g., second-person pronouns like "you," or a direct question answering a preceding statement) makes it impossible to mistake for a single utterance.

{{
  "reasoning": "Provide a brief, explanation for your score.",
  "score": "integer"
}}
"""
\end{Verbatim}
\end{tcolorbox}
\end{center}
\label{app:refine-prompt-1}
\caption{The prompt used to identify and filter semantically ambiguous samples for the Janus-Test using a 5-point Likert scale.}

\end{figure*}

%% file: appendix/prompt_RSF.tex
\begin{figure*}[ht]
\begin{center}
\begin{tcolorbox}[colback=white, coltext=black, title=\textbf{Prompt for Evaluating Response Strategy Following (RSF)}
]
\scriptsize
\begin{Verbatim}[breaklines=true]
# ROLE
Your role is to critically assess the assistant’s responses and determine whether they correctly follow the predefined answer strategy in the situation between the primary speaker and an interrupting third party.
# Evaluation Task
Your task is to decide if the VA followed the strategy in a predefined way when handling a primary speaker request with an immediate third-party interruption. The interruption type is explicitly given: either **Ignorable** or **Actionable**.
## CONTEXT FOR EVALUATION
Primary Speaker Utterance: {user_utterance}
Third-party Interruption after the primary speaker utterance: {interference_utterance}
Interruption Type: {interruption_type}
Assistant's Response: {user_continuous_inference_result}
## REQUIRED STRATEGY
- If the case of Ignorable, the response should ignore the interruption part entirely and respond only to the Primary speaker's original request.
- If the case of Actionable, the response should treat the interruption as helpful but unverified input, incorporate it into a proactive proposal, and explicitly ask the Primary Speaker for final confirmation.
## Scoring Rules
- 1: The assistant corretly follows the required strategy for the given interruption type.
- 0: The assistant does not follow the required strategy, but responds in a different way.
### OUTPUT INSTRUCTIONS
You must strictly adhere to the following format. Your final output must be a single, valid JSON object. Do not add any text before or after the JSON object.
The rationale must be a brief justification for the score, **concise and within 3 sentences.**
**JSON Format:**
```json
{{
  "rationale": "[Your brief justification for the score]",
  "score": [A single integer 0 or 1]
}}
\end{Verbatim}
\end{tcolorbox}
\end{center}
\label{app:refine-prompt-1}
\caption{The prompt used to evaluate how well a trained model's response adheres to the predefined $\pi_A$ and $\pi_I$ strategies.}

\end{figure*}

%% file: appendix/prompt_OH_TPI.tex
\begin{figure*}[ht]
\begin{center}
\begin{tcolorbox}[colback=white, coltext=black, title=\textbf{Prompt for Evaluating Overall Helpfulness on TPI-Eval}
]
\scriptsize
\begin{Verbatim}[breaklines=true]
# ROLE
You are an expert AI Quality Rater specializing in conversational flow and complex interaction(sepecially third-party interruption) scenarios.

# GOAL
Your primary goal is to evaluate the quality of a voice assistant's response within a simulated conversational scenario involving a third-party interruption.

# EVALUATION TASK
Your task is to assess whether the voice assistant model correctly interprets and responds to a third-party interruption spoken by a different person. The evaluation hinges on how the model processes the primary user's speech followed by the interruption from another voice.
You will rate the naturalness, coherence, and helpfulness of the assistant's final response. The score should reflect how well the model understood the conversational context, specifically whether it recognized the interruption as a separate input from a different speaker.
Please note: The response being evaluated may or may not be the one of capable of distinguishing between different speakers. So you should scrunitize the response of voice assistant.
- A high score should be given if the assistant's response demonstrates an understanding that the interruption came from a different individual and addresses the situation appropriately.
- A low score should be given if the response indicates that the model failed to distinguish between the speakers, instead processing the primary utterance and the interruption as a single, continuous statement from the same person.

### CONTEXT FOR EVALUATION
* **Primary Speaker Utterance:** `{user_utterance}`
* **Third-party Interruption after primary speaker utterance:** `{interference_utterance}`
* **Voice Assistant's Response:** `{Inference_result}`

### SCORING GUIDELINES (1-5)

- 1 point: Critical Failure. The model fails to distinguish between the speakers and combines their utterances. This results in a nonsensical, wrong, or completely irrelevant response that doesn't match the intent of both speakers.
- 2 points: Degraded Quality. The model just interpret query by just combining the two utterances, leading to a response that is clearly out of context or awkward. It is obvious that the model misunderstood the multi-speaker situation as a single speaker situation, significantly degrading the quality of the conversation.
- 3 points: Evasive or Incomplete Response. The model fails to handle the interruption gracefully. Instead of directly addressing or perfectly ignoring the interruption (e.g., handling interruption by distinguishing timbre), it produces a generic "safe" response. This type of answer is not strictly incorrect, but it avoids the core difficulty of the request, leaving the impression that the model is luckily hedging rather than providing a substantive solution.
- 4 points: Successful Handling. The model correctly understands the conversational context and makes a reasonable decision. It either (A) addresses the most relevant user intent by incorporating the interruption or (B) correctly determines the interruption is irrelevant and responds only to the primary speaker. The response is helpful and accurate, though it may be slightly less natural or polished (e.g., minor awkwardness in phrasing or flow).
- 5 points: Excellent and Natural Handling. The model flawlessly discerns the intent and relationship between both utterances. It makes the ideal decision to either (A) gracefully integrate the interruption, providing a perfectly tailored response, or (B) seamlessly disregard the interruption as noise and continue the main thread. The final response is not only accurate and helpful but also exceptionally natural and fluid.
---

### OUTPUT INSTRUCTIONS
You must strictly adhere to the following format. Your final output must be a single, valid JSON object. Do not add any text before or after the JSON object.
The rationale must be a brief justification for the score, **concise and within 3 sentences.**

**JSON Format:**
```json
{{
  "rationale": "[Your brief justification for the score]",
  "score": [A single integer from 1 to 5]
}}
\end{Verbatim}
\end{tcolorbox}
\end{center}
\label{app:refine-prompt-1}
\caption{The prompt used to measure the overall helpfulness of model responses on the TPI-Eval set.}

\end{figure*}

%% file: appendix/prompt_OH_Janus.tex
\begin{figure*}[ht]
\begin{center}
\begin{tcolorbox}[colback=white, coltext=black, title=\textbf{Prompt for Evaluating Overall Helpfulness on Janus-Test}
]
\scriptsize
\begin{Verbatim}[breaklines=true]
# ROLE
You are an expert AI Quality Rater specializing in conversational Artificial Intelligence, specifically focusing on Multi-Party Situations and Third-Party Interruptions (TPI).

# GOAL
Your goal is to evaluate the response of a voice assistant in a scenario where a "Third-party Interruption" occurs immediately after the "Primary User's" utterance. You must determine if the assistant correctly distinguishes between the speakers and handles the situation appropriately.

# EVALUATION TASK
You must analyze the "Voice Assistant's Response" based on the following criteria:

1. **Speaker Distinction (Crucial):** Did the assistant recognize that the input consists of two distinct speakers (Primary User vs. Interrupter)? Or did it mistakenly process the entire input as a single continuous sentence from one person?
2. **Primary User Prioritization:** Did the assistant prioritize the intent of the Primary User?
3. **Appropriate Handling Strategy:**
   - If the interruption is irrelevant/noise: The assistant should **ignore** it and answer the primary user.
   - If the interruption is relevant/urgent: The assistant should briefly acknowledge it or incorporate it logically, but never at the expense of confusing the primary user's intent.

**CRITICAL INSTRUCTION:**
- If the model merges the two utterances into one context (treating them as a single speaker), you must give the **LOWEST score**.
- If the model correctly ignores the interruption to focus on the primary user, this is considered a **HIGH quality** response.

---

### CONTEXT FOR EVALUATION
* **Primary Speaker Utterance:** {user_utterance}
* **Third-party Interruption (different speaker):** {interference_utterance}
* **Voice Assistant's Response:** {Inference_result}

---

### SCORING GUIDELINES (1-5)

**1 Point: Critical Failure (Context Merging)**
The model completely fails to distinguish between speakers. It interprets the Primary Utterance and the Interruption as a single, continuous statement from one person. The response is nonsensical, answers a combined query that doesn't exist, or addresses the interruption as if it were part of the user's request.

**2 Points: Significant Confusion**
The model seems confused by the input. It may generate a generic error or a response that tries to address both utterances clunkily, showing no clear understanding that there are two distinct sources. The primary user's intent is lost or heavily distorted.

**3 Points: Safe but Evasive (Hedging)**
The response is not strictly wrong but fails to demonstrate "intelligence." It gives a generic, safe answer that avoids the complexity of the interruption. It does not clearly show whether it distinguished the speakers, leaving the user unsure if the assistant understood the situation.

**4 Points: Successful Handling**
The model correctly identifies the primary user's intent. It handles the interruption reasonably well—either by ignoring it (if it's noise) or addressing the main query despite the distraction. The response is accurate and helpful, though phrasing might be slightly unnatural or rigid.

**5 Points: Excellent & Natural Handling**
The model flawlessly distinguishes the two speakers. It makes the **ideal decision**:
- (A) If the interruption is irrelevant, it strictly ignores it and serves the primary user perfectly.
- (B) If the interruption is relevant, it handles it gracefully without losing focus on the primary user.
The response makes the user feel that the assistant clearly "heard" who said what.

---

### OUTPUT INSTRUCTIONS
Provide your evaluation in the following JSON format.
The rationale must be concise (max 3 sentences) and explicitly state whether the model successfully distinguished the two speakers.

**JSON Format:**
```json
{{
  "rationale": "[Your brief justification for the score]",
  "score": [A single integer from 1 to 5]
}}
\end{Verbatim}
\end{tcolorbox}
\end{center}
\label{app:refine-prompt-1}
\caption{The prompt used to measure the overall helpfulness of model responses on the Janus-Test set.}
\end{figure*}

%% file: appendix/prompt_realworld_filtering.tex
\begin{figure*}[ht]
\begin{center}
\begin{tcolorbox}[colback=white, coltext=black, title=\textbf{Prompt for filtering the samples of real world benchmark}
]
\scriptsize
\begin{Verbatim}[breaklines=true]
You are an expert data annotator.
Your task is to analyze the following **TWO consecutive turns** from a dialogue and rate the extent to which they fit a specific "Interruption of a Query" pattern using a **5-point Likert scale**.

**Input:**
- Turn 1 (Speaker A)
- Turn 2 (Speaker B)

**Criteria to Evaluate (Strict Definition):**

1.  **Turn 1 (VA-Compatible Query):** Speaker A is asking something that a **Voice Assistant (or text-based voice assistant)** could help with.
    * **Scope:** This includes requests that a text-based assistant could deal with, such as **knowledge, facts, definitions, explanations**, etc.
    * Even in a casual conversation, the content should be something an AI could reasonably answer (e.g., "What year did that movie come out?", "What implies a rhetorical question?").

2.  **Turn 2 (Interruption):** Speaker B interrupts Speaker A.
    * Speaker B starts talking before Speaker A finishes (barge-in), OR immediately cuts them off.

**Scoring Instruction:**
- Assign a **Score (1-5)** representing how well the dialogue pair matches the strict criteria above.
- **5:** Strong Agreement (Perfect match; Valid VA/Knowledge query AND Clear interruption).
- **1:** Strong Disagreement (No match).
- Use intermediate scores (2, 3, 4) to reflect the degree of certainty.

**Output Format:**
Respond in strict **JSON** format only. Do not include markdown blocks.

{
  "score": <int, 1-5>,
  "reasoning": "Briefly explain if Turn 1 fits the 'knowledge/fact query' definition and if Turn 2 is an interruption."
}

**Dialogue Pair to Analyze:**
Turn 1 ({spk1}): {txt1}
Turn 2 ({spk2}): {txt2}
\end{Verbatim}
\end{tcolorbox}
\end{center}

\caption{The prompt used to filter real world benchmark samples.}
\label{fig:refining_real_world}
\end{figure*}

%% file: custom.bib
@misc{xu2025qwen3omnitechnicalreport,
      title={Qwen3-Omni Technical Report}, 
      author={Jin Xu and Zhifang Guo and Hangrui Hu and Yunfei Chu and Xiong Wang and Jinzheng He and Yuxuan Wang and Xian Shi and Ting He and Xinfa Zhu and Yuanjun Lv and Yongqi Wang and Dake Guo and He Wang and Linhan Ma and Pei Zhang and Xinyu Zhang and Hongkun Hao and Zishan Guo and Baosong Yang and Bin Zhang and Ziyang Ma and Xipin Wei and Shuai Bai and Keqin Chen and Xuejing Liu and Peng Wang and Mingkun Yang and Dayiheng Liu and Xingzhang Ren and Bo Zheng and Rui Men and Fan Zhou and Bowen Yu and Jianxin Yang and Le Yu and Jingren Zhou and Junyang Lin},
      year={2025},
      eprint={2509.17765},
      archivePrefix={arXiv},
      primaryClass={cs.CL},
      url={https://arxiv.org/abs/2509.17765}, 
}

@article{chen2024slam,
  title={SLAM-Omni: Timbre-Controllable Voice Interaction System with Single-Stage Training},
  author={Chen, Wenxi and Ma, Ziyang and Yan, Ruiqi and Liang, Yuzhe and Li, Xiquan and Xu, Ruiyang and Niu, Zhikang and Zhu, Yanqiao and Yang, Yifan and Liu, Zhanxun and others},
  journal={arXiv preprint arXiv:2412.15649},
  year={2024}
}

@misc{xie2024miniomnilanguagemodelshear,
      title={Mini-Omni: Language Models Can Hear, Talk While Thinking in Streaming}, 
      author={Zhifei Xie and Changqiao Wu},
      year={2024},
      eprint={2408.16725},
      archivePrefix={arXiv},
      primaryClass={cs.AI},
      url={https://arxiv.org/abs/2408.16725}, 
}

@misc{wang2024friendsmmcdatasetmultimodalmultiparty,
      title={Friends-MMC: A Dataset for Multi-modal Multi-party Conversation Understanding}, 
      author={Yueqian Wang and Xiaojun Meng and Yuxuan Wang and Jianxin Liang and Qun Liu and Dongyan Zhao},
      year={2024},
      eprint={2412.17295},
      archivePrefix={arXiv},
      primaryClass={cs.CL},
      url={https://arxiv.org/abs/2412.17295}, 
}

@inproceedings{10.1007/11677482_3,
author = {Carletta, Jean and Ashby, Simone and Bourban, Sebastien and Flynn, Mike and Guillemot, Mael and Hain, Thomas and Kadlec, Jaroslav and Karaiskos, Vasilis and Kraaij, Wessel and Kronenthal, Melissa and Lathoud, Guillaume and Lincoln, Mike and Lisowska, Agnes and McCowan, Iain and Post, Wilfried and Reidsma, Dennis and Wellner, Pierre},
title = {The AMI meeting corpus: a pre-announcement},
year = {2005},
isbn = {3540325492},
publisher = {Springer-Verlag},
address = {Berlin, Heidelberg},
url = {https://doi.org/10.1007/11677482_3},
doi = {10.1007/11677482_3},
booktitle = {Proceedings of the Second International Conference on Machine Learning for Multimodal Interaction},
pages = {28–39},
numpages = {12},
location = {Edinburgh, UK},
series = {MLMI'05}
}

@misc{openai2025chatgpt,
  author       = {OpenAI},
  title        = {ChatGPT},
  year         = {2025},
  howpublished = {\url{https://chat.openai.com}},
  note         = {Large language model interface. Accessed: 2025-09-22}
}

@misc{yang2025qwen3technicalreport,
      title={Qwen3 Technical Report}, 
      author={An Yang and Anfeng Li and Baosong Yang and Beichen Zhang and Binyuan Hui and Bo Zheng and Bowen Yu and Chang Gao and Chengen Huang and Chenxu Lv and Chujie Zheng and Dayiheng Liu and Fan Zhou and Fei Huang and Feng Hu and Hao Ge and Haoran Wei and Huan Lin and Jialong Tang and Jian Yang and Jianhong Tu and Jianwei Zhang and Jianxin Yang and Jiaxi Yang and Jing Zhou and Jingren Zhou and Junyang Lin and Kai Dang and Keqin Bao and Kexin Yang and Le Yu and Lianghao Deng and Mei Li and Mingfeng Xue and Mingze Li and Pei Zhang and Peng Wang and Qin Zhu and Rui Men and Ruize Gao and Shixuan Liu and Shuang Luo and Tianhao Li and Tianyi Tang and Wenbiao Yin and Xingzhang Ren and Xinyu Wang and Xinyu Zhang and Xuancheng Ren and Yang Fan and Yang Su and Yichang Zhang and Yinger Zhang and Yu Wan and Yuqiong Liu and Zekun Wang and Zeyu Cui and Zhenru Zhang and Zhipeng Zhou and Zihan Qiu},
      year={2025},
      eprint={2505.09388},
      archivePrefix={arXiv},
      primaryClass={cs.CL},
      url={https://arxiv.org/abs/2505.09388}, 
}

@inproceedings{Pratap_2020,
  title     = {{MLS}: A Large-Scale Multilingual Dataset for Speech Research},
  author    = {Pratap, Vineel and Xu, Qiantong and Sriram, Anuroop and Synnaeve, Gabriel and Collobert, Ronan},
  booktitle = {Interspeech 2020},
  year      = {2020},
  month     = {10},
  pages     = {2757--2761},
  publisher = {ISCA},
  doi       = {10.21437/interspeech.2020-2826},
  url       = {http://dx.doi.org/10.21437/Interspeech.2020-2826}
}

@misc{zhang2025omniflattenendtoendgptmodel,
      title={OmniFlatten: An End-to-end GPT Model for Seamless Voice Conversation}, 
      author={Qinglin Zhang and Luyao Cheng and Chong Deng and Qian Chen and Wen Wang and Siqi Zheng and Jiaqing Liu and Hai Yu and Chaohong Tan and Zhihao Du and Shiliang Zhang},
      year={2025},
      eprint={2410.17799},
      archivePrefix={arXiv},
      primaryClass={cs.CL},
      url={https://arxiv.org/abs/2410.17799}, 
}

@inproceedings{stacey-etal-2024-lucid,
    title = "{LUCID}: {LLM}-Generated Utterances for Complex and Interesting Dialogues",
    author = "Stacey, Joe  and
      Cheng, Jianpeng  and
      Torr, John  and
      Guigue, Tristan  and
      Driesen, Joris  and
      Coca, Alexandru  and
      Gaynor, Mark  and
      Johannsen, Anders",
    editor = "Cao, Yang (Trista)  and
      Papadimitriou, Isabel  and
      Ovalle, Anaelia  and
      Zampieri, Marcos  and
      Ferraro, Francis  and
      Swayamdipta, Swabha",
    booktitle = "Proceedings of the 2024 Conference of the North American Chapter of the Association for Computational Linguistics: Human Language Technologies (Volume 4: Student Research Workshop)",
    month = jun,
    year = "2024",
    address = "Mexico City, Mexico",
    publisher = "Association for Computational Linguistics",
    url = "https://aclanthology.org/2024.naacl-srw.8/",
    doi = "10.18653/v1/2024.naacl-srw.8",
    pages = "56--74"
}

@misc{xu2025qwen25omnitechnicalreport,
      title={Qwen2.5-Omni Technical Report}, 
      author={Jin Xu and Zhifang Guo and Jinzheng He and Hangrui Hu and Ting He and Shuai Bai and Keqin Chen and Jialin Wang and Yang Fan and Kai Dang and Bin Zhang and Xiong Wang and Yunfei Chu and Junyang Lin},
      year={2025},
      eprint={2503.20215},
      archivePrefix={arXiv},
      primaryClass={cs.CL},
      url={https://arxiv.org/abs/2503.20215}, 
}

@misc{chatterboxtts2025,
  author       = {{Resemble AI}},
  title        = {{Chatterbox-TTS}},
  year         = {2025},
  howpublished = {\url{https://github.com/resemble-ai/chatterbox}},
  note         = {GitHub repository}
}

@inproceedings{
cheng2025voxdialogue,
title={VoxDialogue: Can Spoken Dialogue Systems Understand Information Beyond Words?},
author={Xize Cheng and Ruofan Hu and Xiaoda Yang and Jingyu Lu and Dongjie Fu and Zehan Wang and Shengpeng Ji and Rongjie Huang and Boyang Zhang and Tao Jin and Zhou Zhao},
booktitle={The Thirteenth International Conference on Learning Representations},
year={2025},
url={https://openreview.net/forum?id=vbmSSIhKAM}
}

@inproceedings{NEURIPS2023_7b16688a,
author = {Si, Shuzheng and Ma, Wentao and Gao, Haoyu and Wu, Yuchuan and Lin, Ting-En and Dai, Yinpei and Li, Hangyu and Yan, Rui and Huang, Fei and Li, Yongbin},
booktitle = {Advances in Neural Information Processing Systems},
editor = {A. Oh and T. Neumann and A. Globerson and K. Saenko and M. Hardt and S. Levine},
pages = {39088--39118},
publisher = {Curran Associates, Inc.},
title = {SpokenWOZ: A Large-Scale Speech-Text Benchmark for Spoken Task-Oriented Dialogue Agents},
url = {https://proceedings.neurips.cc/paper_files/paper/2023/file/ 7b16688a2b053a1b01474ab5c78ce662-Paper-Datasets_and_Benchmarks.pdf},
volume = {36},
year={2023}}

@misc{koudounas2025deepdialoguemultiturnemotionallyrichspoken,
      title={DeepDialogue: A Multi-Turn Emotionally-Rich Spoken Dialogue Dataset}, 
      author={Alkis Koudounas and Moreno La Quatra and Elena Baralis},
      year={2025},
      eprint={2505.19978},
      archivePrefix={arXiv},
      primaryClass={cs.CL},
      url={https://arxiv.org/abs/2505.19978}, 

}

@inproceedings{Wang_2024,
  title     = {{DiarizationLM}: Speaker Diarization Post-Processing with Large Language Models},
  author    = {Wang, Quan and Huang, Yiling and Zhao, Guanlong and Clark, Evan and Xia, Wei and Liao, Hank},
  booktitle = {Interspeech 2024},
  year      = {2024},
  month     = {9},
  pages     = {3754--3758},
  publisher = {ISCA},
  doi       = {10.21437/interspeech.2024-209},
  url       = {http://dx.doi.org/10.21437/Interspeech.2024-209}
}

@misc{he2025surveyendtoendmultispeakerautomatic,
      title={Survey of End-to-End Multi-Speaker Automatic Speech Recognition for Monaural Audio}, 
      author={Xinlu He and Jacob Whitehill},
      year={2025},
      eprint={2505.10975},
      archivePrefix={arXiv},
      primaryClass={cs.CL},
      url={https://arxiv.org/abs/2505.10975}, 
}

@misc{yin2025speakerlmendtoendversatilespeaker,
      title={SpeakerLM: End-to-End Versatile Speaker Diarization and Recognition with Multimodal Large Language Models}, 
      author={Han Yin and Yafeng Chen and Chong Deng and Luyao Cheng and Hui Wang and Chao-Hong Tan and Qian Chen and Wen Wang and Xiangang Li},
      year={2025},
      eprint={2508.06372},
      archivePrefix={arXiv},
      primaryClass={cs.SD},
      url={https://arxiv.org/abs/2508.06372}, 
}

@misc{lin2025diarizationawaremultispeakerautomaticspeech,
      title={Diarization-Aware Multi-Speaker Automatic Speech Recognition via Large Language Models}, 
      author={Yuke Lin and Ming Cheng and Ze Li and Beilong Tang and Ming Li},
      year={2025},
      eprint={2506.05796},
      archivePrefix={arXiv},
      primaryClass={eess.AS},
      url={https://arxiv.org/abs/2506.05796}, 
}

@misc{wang2025speakertargetingselfspeakeradaptation,
      title={Speaker Targeting via Self-Speaker Adaptation for Multi-talker ASR}, 
      author={Weiqing Wang and Taejin Park and Ivan Medennikov and Jinhan Wang and Kunal Dhawan and He Huang and Nithin Rao Koluguri and Jagadeesh Balam and Boris Ginsburg},
      year={2025},
      eprint={2506.22646},
      archivePrefix={arXiv},
      primaryClass={eess.AS},
      url={https://arxiv.org/abs/2506.22646}, 
}

@misc{yan2025urobenchcomprehensiveevaluationendtoend,
      title={URO-Bench: Towards Comprehensive Evaluation for End-to-End Spoken Dialogue Models}, 
      author={Ruiqi Yan and Xiquan Li and Wenxi Chen and Zhikang Niu and Chen Yang and Ziyang Ma and Kai Yu and Xie Chen},
      year={2025},
      eprint={2502.17810},
      archivePrefix={arXiv},
      primaryClass={cs.CL},
      url={https://arxiv.org/abs/2502.17810}, 
}

@misc{lee2023dailytalkspokendialoguedataset,
      title={DailyTalk: Spoken Dialogue Dataset for Conversational Text-to-Speech}, 
      author={Keon Lee and Kyumin Park and Daeyoung Kim},
      year={2023},
      eprint={2207.01063},
      archivePrefix={arXiv},
      primaryClass={eess.AS},
      url={https://arxiv.org/abs/2207.01063}, 
}

@misc{xu2025targetspeakerextractioncomparing,
      title={Target Speaker Extraction through Comparing Noisy Positive and Negative Audio Enrollments}, 
      author={Shitong Xu and Yiyuan Yang and Niki Trigoni and Andrew Markham},
      year={2025},
      eprint={2502.16611},
      archivePrefix={arXiv},
      primaryClass={cs.SD},
      url={https://arxiv.org/abs/2502.16611}, 
}

@misc{meng2025largelanguagemodeltranscribe,
      title={Large Language Model Can Transcribe Speech in Multi-Talker Scenarios with Versatile Instructions}, 
      author={Lingwei Meng and Shujie Hu and Jiawen Kang and Zhaoqing Li and Yuejiao Wang and Wenxuan Wu and Xixin Wu and Xunying Liu and Helen Meng},
      year={2025},
      eprint={2409.08596},
      archivePrefix={arXiv},
      primaryClass={cs.CL},
      url={https://arxiv.org/abs/2409.08596}, 
}

@misc{chen2024voicebenchbenchmarkingllmbasedvoice,
      title={VoiceBench: Benchmarking LLM-Based Voice Assistants}, 
      author={Yiming Chen and Xianghu Yue and Chen Zhang and Xiaoxue Gao and Robby T. Tan and Haizhou Li},
      year={2024},
      eprint={2410.17196},
      archivePrefix={arXiv},
      primaryClass={cs.CL},
      url={https://arxiv.org/abs/2410.17196}, 
}

@article{xie2022proactivity,
  title     = {Proactivity or passivity? An investigation of the effect of service robots’ proactive behaviour on customer co-creation intention},
  author    = {Xie, Lishan and Liu, Canmian and Li, Dongmei},
  journal   = {International Journal of Hospitality Management},
  volume    = {106},
  pages     = {103271},
  year      = {2022},
  issn      = {0278-4319},
  doi       = {10.1016/j.ijhm.2022.103271},
  url       = {https://www.sciencedirect.com/science/article/pii/S0278431922001335}
}

@misc{ao2025sdevalbenchmarkdatasetspoken,
      title={SD-Eval: A Benchmark Dataset for Spoken Dialogue Understanding Beyond Words}, 
      author={Junyi Ao and Yuancheng Wang and Xiaohai Tian and Dekun Chen and Jun Zhang and Lu Lu and Yuxuan Wang and Haizhou Li and Zhizheng Wu},
      year={2025},
      eprint={2406.13340},
      archivePrefix={arXiv},
      primaryClass={cs.CL},
      url={https://arxiv.org/abs/2406.13340}, 
}

@article{10.1145/3369807,
author = {Tabassum, Madiha and Kosi\'{n}ski, Tomasz and Frik, Alisa and Malkin, Nathan and Wijesekera, Primal and Egelman, Serge and Lipford, Heather Richter},
title = {Investigating Users' Preferences and Expectations for Always-Listening Voice Assistants},
year = {2020},
issue_date = {December 2019},
publisher = {Association for Computing Machinery},
address = {New York, NY, USA},
volume = {3},
number = {4},
url = {https://doi.org/10.1145/3369807},
doi = {10.1145/3369807},
journal = {Proc. ACM Interact. Mob. Wearable Ubiquitous Technol.},
month = sep,
articleno = {153},
numpages = {23}
}

@misc{koishigarina2025clipbehaveslikebagofwords,
      title={CLIP Behaves like a Bag-of-Words Model Cross-modally but not Uni-modally}, 
      author={Darina Koishigarina and Arnas Uselis and Seong Joon Oh},
      year={2025},
      eprint={2502.03566},
      archivePrefix={arXiv},
      primaryClass={cs.CV},
      url={https://arxiv.org/abs/2502.03566}, 
}

@misc{radford2022whisper,
  doi = {10.48550/ARXIV.2212.04356},
  url = {https://arxiv.org/abs/2212.04356},
  author = {Radford, Alec and Kim, Jong Wook and Xu, Tao and Brockman, Greg and McLeavey, Christine and Sutskever, Ilya},
  title = {Robust Speech Recognition via Large-Scale Weak Supervision},
  publisher = {arXiv},
  year = {2022},
  copyright = {arXiv.org perpetual, non-exclusive license}
}

@misc{yuksekgonul2023visionlanguagemodelsbehavelike,
      title={When and why vision-language models behave like bags-of-words, and what to do about it?}, 
      author={Mert Yuksekgonul and Federico Bianchi and Pratyusha Kalluri and Dan Jurafsky and James Zou},
      year={2023},
      eprint={2210.01936},
      archivePrefix={arXiv},
      primaryClass={cs.CV},
      url={https://arxiv.org/abs/2210.01936}, 
}

@misc{cao2025interruptionhandlingconversationalrobots,
      title={Interruption Handling for Conversational Robots}, 
      author={Shiye Cao and Jiwon Moon and Amama Mahmood and Victor Nikhil Antony and Ziang Xiao and Anqi Liu and Chien-Ming Huang},
      year={2025},
      eprint={2501.01568},
      archivePrefix={arXiv},
      primaryClass={cs.HC},
      url={https://arxiv.org/abs/2501.01568}, 
}

@misc{long2025vitaaudiofastinterleavedcrossmodal,
      title={VITA-Audio: Fast Interleaved Cross-Modal Token Generation for Efficient Large Speech-Language Model}, 
      author={Zuwei Long and Yunhang Shen and Chaoyou Fu and Heting Gao and Lijiang Li and Peixian Chen and Mengdan Zhang and Hang Shao and Jian Li and Jinlong Peng and Haoyu Cao and Ke Li and Rongrong Ji and Xing Sun},
      year={2025},
      eprint={2505.03739},
      archivePrefix={arXiv},
      primaryClass={cs.CL},
      url={https://arxiv.org/abs/2505.03739}, 
}

@article{Goldberg1990InterruptingTD,
  title={Interrupting the discourse on interruptions},
  author={Julia Goldberg},
  journal={Journal of Pragmatics},
  year={1990},
  volume={14},
  pages={883-903},
  url={https://api.semanticscholar.org/CorpusID:142727846}
}

@article{Murata1994,
  author    = {Kumiko Murata},
  title     = {Intrusive or co-operative? A cross-cultural study of interruption},
  journal   = {Journal of Pragmatics},
  volume    = {21},
  number    = {4},
  pages     = {385--400},
  year      = {1994},
  issn      = {0378-2166},
  doi       = {10.1016/0378-2166(94)90011-6},
  url       = {https://www.sciencedirect.com/science/article/pii/0378216694900116}
}

@inproceedings{yang-etal-2022-annotating,
    title = "Annotating Interruption in Dyadic Human Interaction",
    author = "Yang, Liu  and
      Achard, Catherine  and
      Pelachaud, Catherine",
    editor = "Calzolari, Nicoletta  and
      B{\'e}chet, Fr{\'e}d{\'e}ric  and
      Blache, Philippe  and
      Choukri, Khalid  and
      Cieri, Christopher  and
      Declerck, Thierry  and
      Goggi, Sara  and
      Isahara, Hitoshi  and
      Maegaard, Bente  and
      Mariani, Joseph  and
      Mazo, H{\'e}l{\`e}ne  and
      Odijk, Jan  and
      Piperidis, Stelios",
    booktitle = "Proceedings of the Thirteenth Language Resources and Evaluation Conference",
    month = jun,
    year = "2022",
    address = "Marseille, France",
    publisher = "European Language Resources Association",
    url = "https://aclanthology.org/2022.lrec-1.245/",
    pages = "2292--2297"
}

@misc{kimiteam2025kimiaudiotechnicalreport,
      title={Kimi-Audio Technical Report}, 
      author={KimiTeam and Ding Ding and Zeqian Ju and Yichong Leng and Songxiang Liu and Tong Liu and Zeyu Shang and Kai Shen and Wei Song and Xu Tan and Heyi Tang and Zhengtao Wang and Chu Wei and Yifei Xin and Xinran Xu and Jianwei Yu and Yutao Zhang and Xinyu Zhou and Y. Charles and Jun Chen and Yanru Chen and Yulun Du and Weiran He and Zhenxing Hu and Guokun Lai and Qingcheng Li and Yangyang Liu and Weidong Sun and Jianzhou Wang and Yuzhi Wang and Yuefeng Wu and Yuxin Wu and Dongchao Yang and Hao Yang and Ying Yang and Zhilin Yang and Aoxiong Yin and Ruibin Yuan and Yutong Zhang and Zaida Zhou},
      year={2025},
      eprint={2504.18425},
      archivePrefix={arXiv},
      primaryClass={eess.AS},
      url={https://arxiv.org/abs/2504.18425}, 
}

@misc{kim2024paralinguisticsawarespeechempoweredlargelanguage,
      title={Paralinguistics-Aware Speech-Empowered Large Language Models for Natural Conversation}, 
      author={Heeseung Kim and Soonshin Seo and Kyeongseok Jeong and Ohsung Kwon and Soyoon Kim and Jungwhan Kim and Jaehong Lee and Eunwoo Song and Myungwoo Oh and Jung-Woo Ha and Sungroh Yoon and Kang Min Yoo},
      year={2024},
      eprint={2402.05706},
      archivePrefix={arXiv},
      primaryClass={cs.CL},
      url={https://arxiv.org/abs/2402.05706}, 
}

@misc{wu2025stepaudio2technicalreport,
      title={Step-Audio 2 Technical Report}, 
      author={Boyong Wu and Chao Yan and Chen Hu and Cheng Yi and Chengli Feng and Fei Tian and Feiyu Shen and Gang Yu and Haoyang Zhang and Jingbei Li and Mingrui Chen and Peng Liu and Wang You and Xiangyu Tony Zhang and Xingyuan Li and Xuerui Yang and Yayue Deng and Yechang Huang and Yuxin Li and Yuxin Zhang and Zhao You and Brian Li and Changyi Wan and Hanpeng Hu and Jiangjie Zhen and Siyu Chen and Song Yuan and Xuelin Zhang and Yimin Jiang and Yu Zhou and Yuxiang Yang and Bingxin Li and Buyun Ma and Changhe Song and Dongqing Pang and Guoqiang Hu and Haiyang Sun and Kang An and Na Wang and Shuli Gao and Wei Ji and Wen Li and Wen Sun and Xuan Wen and Yong Ren and Yuankai Ma and Yufan Lu and Bin Wang and Bo Li and Changxin Miao and Che Liu and Chen Xu and Dapeng Shi and Dingyuan Hu and Donghang Wu and Enle Liu and Guanzhe Huang and Gulin Yan and Han Zhang and Hao Nie and Haonan Jia and Hongyu Zhou and Jianjian Sun and Jiaoren Wu and Jie Wu and Jie Yang and Jin Yang and Junzhe Lin and Kaixiang Li and Lei Yang and Liying Shi and Li Zhou and Longlong Gu and Ming Li and Mingliang Li and Mingxiao Li and Nan Wu and Qi Han and Qinyuan Tan and Shaoliang Pang and Shengjie Fan and Siqi Liu and Tiancheng Cao and Wanying Lu and Wenqing He and Wuxun Xie and Xu Zhao and Xueqi Li and Yanbo Yu and Yang Yang and Yi Liu and Yifan Lu and Yilei Wang and Yuanhao Ding and Yuanwei Liang and Yuanwei Lu and Yuchu Luo and Yuhe Yin and Yumeng Zhan and Yuxiang Zhang and Zidong Yang and Zixin Zhang and Binxing Jiao and Daxin Jiang and Heung-Yeung Shum and Jiansheng Chen and Jing Li and Xiangyu Zhang and Yibo Zhu},
      year={2025},
      eprint={2507.16632},
      archivePrefix={arXiv},
      primaryClass={cs.CL},
      url={https://arxiv.org/abs/2507.16632}, 
}

@article{LEVELT198341,
	author = {Willem J.M. Levelt},
	doi = {https://doi.org/10.1016/0010-0277(83)90026-4},
	issn = {0010-0277},
	journal = {Cognition},
	number = {1},
	pages = {41-104},
	title = {Monitoring and self-repair in speech},
	url = {https://www.sciencedirect.com/science/article/pii/0010027783900264},
	volume = {14},
	year = {1983}}

@misc{wang2025msubenchunderstandingconversationalmultitalker,
      title={MSU-Bench: Towards Understanding the Conversational Multi-talker Scenarios}, 
      author={Shuai Wang and Zhaokai Sun and Zhennan Lin and Chengyou Wang and Zhou Pan and Lei Xie},
      year={2025},
      eprint={2508.08155},
      archivePrefix={arXiv},
      primaryClass={eess.AS},
      url={https://arxiv.org/abs/2508.08155}, 
}

@misc{liu2024toadtaskorientedautomaticdialogs,
      title={TOAD: Task-Oriented Automatic Dialogs with Diverse Response Styles}, 
      author={Yinhong Liu and Yimai Fang and David Vandyke and Nigel Collier},
      year={2024},
      eprint={2402.10137},
      archivePrefix={arXiv},
      primaryClass={cs.CL},
      url={https://arxiv.org/abs/2402.10137}, 
}

@misc{goel2023prestomultilingualdatasetparsing,
      title={PRESTO: A Multilingual Dataset for Parsing Realistic Task-Oriented Dialogs}, 
      author={Rahul Goel and Waleed Ammar and Aditya Gupta and Siddharth Vashishtha and Motoki Sano and Faiz Surani and Max Chang and HyunJeong Choe and David Greene and Kyle He and Rattima Nitisaroj and Anna Trukhina and Shachi Paul and Pararth Shah and Rushin Shah and Zhou Yu},
      year={2023},
      eprint={2303.08954},
      archivePrefix={arXiv},
      primaryClass={cs.CL},
      url={https://arxiv.org/abs/2303.08954}, 
}

@misc{gu2025surveyllmasajudge,
      title={A Survey on LLM-as-a-Judge}, 
      author={Jiawei Gu and Xuhui Jiang and Zhichao Shi and Hexiang Tan and Xuehao Zhai and Chengjin Xu and Wei Li and Yinghan Shen and Shengjie Ma and Honghao Liu and Saizhuo Wang and Kun Zhang and Yuanzhuo Wang and Wen Gao and Lionel Ni and Jian Guo},
      year={2025},
      eprint={2411.15594},
      archivePrefix={arXiv},
      primaryClass={cs.CL},
      url={https://arxiv.org/abs/2411.15594}, 
}

@misc{kwon2025m3sluevaluatingspeakerattributedreasoning,
      title={M3-SLU: Evaluating Speaker-Attributed Reasoning in Multimodal Large Language Models}, 
      author={Yejin Kwon and Taewoo Kang and Hyunsoo Yoon and Changouk Kim},
      year={2025},
      eprint={2510.19358},
      archivePrefix={arXiv},
      primaryClass={cs.CL},
      url={https://arxiv.org/abs/2510.19358}, 
}

@misc{stacey2024lucidllmgeneratedutterancescomplex,
      title={LUCID: LLM-Generated Utterances for Complex and Interesting Dialogues}, 
      author={Joe Stacey and Jianpeng Cheng and John Torr and Tristan Guigue and Joris Driesen and Alexandru Coca and Mark Gaynor and Anders Johannsen},
      year={2024},
      eprint={2403.00462},
      archivePrefix={arXiv},
      primaryClass={cs.CL},
      url={https://arxiv.org/abs/2403.00462}, 
}

@inproceedings{10.1145/3544548.3581152,
author = {Baughan, Amanda and Wang, Xuezhi and Liu, Ariel and Mercurio, Allison and Chen, Jilin and Ma, Xiao},
title = {A Mixed-Methods Approach to Understanding User Trust after Voice Assistant Failures},
year = {2023},
isbn = {9781450394215},
publisher = {Association for Computing Machinery},
address = {New York, NY, USA},
url = {https://doi.org/10.1145/3544548.3581152},
doi = {10.1145/3544548.3581152},
booktitle = {Proceedings of the 2023 CHI Conference on Human Factors in Computing Systems},
articleno = {7},
numpages = {16},
location = {Hamburg, Germany},
series = {CHI '23}
}
